%% file: 0main.tex
\documentclass[journal]{IEEEtran}

\usepackage{amsmath,amssymb,amsfonts}

\usepackage{algorithm}
\usepackage{algorithmic}
\usepackage{xcolor}
\usepackage{graphicx}
\usepackage{epstopdf}

\usepackage{array}
\usepackage{multirow}
\usepackage{booktabs}
\usepackage[table]{xcolor}
\usepackage{rotating}

\usepackage{cite}

\usepackage{textcomp}
\usepackage{stfloats}   
\usepackage{url}
\usepackage{balance}    
\usepackage[font=small]{caption}
\usepackage{subfig}
\usepackage{amsmath}
\usepackage{pgfplots}
\usepgfplotslibrary{colormaps}
\usepackage{xcolor}
\usetikzlibrary{positioning}
\pgfplotsset{compat=1.18}
\usepgfplotslibrary{colormaps}
\usepackage{tikz}
\usetikzlibrary{positioning,patterns,calc}

\usepackage{microtype}

\usepackage[colorlinks,linkcolor=blue,citecolor=blue,urlcolor=black]{hyperref}
\usepackage{makecell}
\usepackage{microtype}
\usepackage{stfloats}

\usepackage{amsmath} 
\usepackage{amssymb}
\usepackage{amsfonts}
\usepackage{dsfont}

\usepackage{adjustbox}

\usepackage{pgfplots}
\pgfplotsset{compat=1.18}
\usepackage{pgfplotstable}
\usetikzlibrary{pgfplots.colormaps}

\usepackage{array}

\newcolumntype{C}[1]{>{\centering\arraybackslash}m{#1}}

\begin{document}


\title{SAiW: Source-Attributable Invisible Watermarking for Proactive Deepfake Defense}

\newcommand{{\titleabbr}}{SAiW}

\author{Bibek Das,~\IEEEmembership{Student Member,~IEEE},
Chandranath Adak,~\IEEEmembership{Senior Member,~IEEE}, 
Soumi~Chattopadhyay,~\IEEEmembership{Senior Member,~IEEE},  
Zahid Akhtar,~\IEEEmembership{Senior Member,~IEEE}, 
Soumya Dutta,~\IEEEmembership{Member,~IEEE} 
\thanks{B. Das and C. Adak are with Dept. of CSE, Indian Institute of Technology Patna, India – 801106. \\
S. Chattopadhyay is with Dept. of CSE, Indian Institute of Technology Indore, India – 453552. \\
Z. Akhtar is with State University of New York Polytechnic Institute, NY, USA – 13502. \\ 
S. Dutta is with Dept. of CSE, Indian Institute of Technology Kanpur, India – 208016. \\ 
Corresponding author: Chandranath Adak (chandranath@iitp.ac.in) 
}
}

\markboth{B.~Das \MakeLowercase{\textit{et al.}}: {{\titleabbr}}}%
{}

\maketitle

\input{0abstract}

\input{1intro_CA}

\input{2related_CA}

\input{3method_CA}

\input{5result_CA}

\balance 

\input{6Conclusion}

\bibliographystyle{IEEEtran}
\bibliography{ref}

\section{Supplementary Appendix}

Appendices A, B, C, D, and E can be found in \href{https://github.com/bibek-cse/SAiW}{https://github.com/bibek-cse/SAiW}.

\end{document}

%% file: 0abstract.tex
\begin{abstract}
Deepfakes generated by modern generative models pose a serious threat to information integrity, digital identity, and public trust. Existing detection methods are largely reactive, attempting to identify manipulations after they occur and often failing to generalize across evolving generation techniques. This motivates the need for proactive mechanisms that secure media authenticity at the time of creation. In this work, we introduce {\titleabbr}, a \textit{Source-Attributed Invisible watermarking Framework} for proactive deepfake defense and media provenance verification. 
Unlike conventional watermarking methods that treat watermark payloads as generic signals, {\titleabbr} formulates watermark embedding as a source-conditioned representation learning problem, where watermark identity encodes the originating source and modulates the embedding process to produce discriminative and traceable signatures. The framework integrates feature-wise linear modulation to inject source identity into the embedding network, enabling scalable multi-source watermark generation. A perceptual guidance module derived from human visual system priors ensures that watermark perturbations remain visually imperceptible while maintaining robustness. In addition, a dual-purpose forensic decoder simultaneously reconstructs the embedded watermark and performs source attribution, providing both automated verification and interpretable forensic evidence. 
Extensive experiments across multiple deepfake datasets demonstrate that {\titleabbr} achieves high perceptual quality while maintaining strong robustness against compression, filtering, noise, geometric transformations, and adversarial perturbations. By binding digital media to its origin through invisible yet verifiable markers, {\titleabbr} enables reliable authentication and source attribution, providing a scalable foundation for proactive deepfake defense and trustworthy media provenance.
\end{abstract}

\begin{IEEEkeywords}
Proactive Deepfake Defense, Invisible Watermarking, Source Attribution, Content Authentication
\end{IEEEkeywords}

%% file: 1intro_CA.tex
\section{Introduction}
\label{Intro}

The rapid advancement of generative models has enabled the creation of highly realistic synthetic media, fundamentally challenging the reliability and authenticity of digital visual content. While these technologies enable creative media generation, they also amplify the threat posed by deepfakes \cite{akhtar2024video}. Deepfakes are AI-generated videos, images, or audio that falsely depict individuals performing actions or making statements they never made. Such synthetic content can facilitate misinformation, impersonation, and fraud, while eroding trust in digital evidence, particularly in domains where media authenticity is critical, including journalism, public communication, and legal proceedings.

Among different forms of synthetic manipulation, facial deepfakes have attracted significant attention in the biometrics and visual forensics communities \cite{NGUYEN2022103525,1282,bibekBS}. The human face, one of the most socially and semantically important biometric traits, can now be manipulated using techniques such as identity swapping, attribute editing, and face reenactment \cite{InsightFace,simswap2020,faceapp,FSGAN}. Although these methods were originally developed for entertainment applications, the widespread availability of pre-trained generative models and consumer applications such as FaceApp \cite{faceapp} enables even non-expert users to generate convincing forged media within minutes. Consequently, deepfakes have become powerful tools for misinformation, political manipulation, harassment, financial fraud, and reputational damage, further undermining public trust in digital media. 
To address these challenges, most existing research focuses on deepfake detection methods that attempt to identify manipulation artifacts after synthetic media has been generated. These approaches typically rely on visual inconsistencies, blending artifacts, or physiological irregularities \cite{Oculi, dfdc, Xception, 9578910}. However, such reactive detection strategies face fundamental limitations. Detection models often struggle to generalize across unseen manipulation techniques, suffer from dataset bias, and remain vulnerable to adversarial attacks. Moreover, the rapid evolution of generative models frequently outpaces the adaptability of artifact-based detection systems \cite{Xception, Anti_Forgery}. These limitations motivate a shift from reactive detection toward proactive mechanisms that secure media authenticity at the time of creation.

Proactive deepfake defense aims to embed verifiable authenticity signals directly into digital content, enabling reliable verification of origin and integrity throughout the media lifecycle. Recent approaches include cryptographic provenance tracking, identity-preserving perturbations, and robust digital watermarking \cite{StegaStamp, ProActive}. In parallel, initiatives such as the 
C2PA \cite{C2PA} 
propose embedding signed metadata within digital assets to maintain traceable provenance records. While such metadata-based mechanisms provide useful provenance information, they remain vulnerable to removal or corruption during platform-specific processing or adversarial manipulation \cite{vpn}. As a result, embedding persistent authenticity markers directly within visual content has emerged as a promising complementary solution. 
Despite these advances, existing watermarking techniques still face challenges in deepfake defense. Many methods focus mainly on payload recovery without modeling source attribution or supporting scalable multi-source watermarking. Moreover, jointly achieving imperceptibility, robustness to real-world distortions, and reliable source identification remains difficult.

The key insight of our work is that watermark identity can act not only as a hidden payload but also as a conditioning signal that guides the embedding process, enabling a unified framework for scalable media provenance attribution. 
In this work, we propose {\titleabbr} for proactive deepfake defense and media provenance verification. Unlike conventional approaches that treat watermark payloads as generic signals, our method formulates watermark embedding as a source-conditioned representation learning problem in which watermark identity encodes the originating source and modulates the embedding process to produce discriminative and traceable signatures. Trusted entities embed source-specific invisible watermarks into media at creation, establishing persistent authenticity markers that remain verifiable under subsequent transformations. The framework integrates source-conditioned feature modulation with perceptual guidance based on human visual system characteristics to ensure imperceptibility while maintaining robustness to compression, filtering, noise, and other real-world distortions. In addition, a dual-purpose forensic decoder simultaneously reconstructs the embedded watermark and performs source attribution, enabling both automated verification and interpretable forensic evidence. By binding digital media to its origin through invisible yet verifiable markers, {\titleabbr} provides a scalable mechanism for proactive media authentication and complements existing deepfake detection methods by enabling reliable source authentication and media provenance. 
The major \textbf{contributions} of this paper are summarized as follows: 

\textit{\textbf{(i) Source-conditioned watermarking formulation:}} 
We reformulate digital watermarking as a source-conditioned representation learning problem, where the watermark identity encodes the originating media source and acts as a conditioning signal for the embedding process. Unlike conventional payload-based designs, this formulation enables a unified model to learn source-aware watermark representations and supports scalable provenance encoding across multiple generative sources.

\textit{\textbf{(ii) Content-adaptive perceptual embedding with identity modulation:}} 
We propose a content-adaptive embedding mechanism that integrates perceptual guidance based on human visual system characteristics with watermark identity-driven feature modulation. By combining luminance adaptation and contrast masking with channel-wise conditioning, the framework dynamically allocates embedding strength to visually tolerant regions while enabling source-specific watermark patterns, thereby improving both imperceptibility and robustness.

\textit{\textbf{(iii) Unified forensic decoding for reconstruction and attribution:}} 
We design a dual-purpose decoding architecture that jointly performs watermark reconstruction and source attribution. The shared representation enables reliable watermark recovery under distortions while learning discriminative source embeddings, providing both automated verification and interpretable forensic evidence for media provenance analysis.

The rest of the paper is organized as follows. 
Section~\ref{Review} reviews related work. 
Section~\ref{method} presents the proposed method. 
Section~\ref{Exp} describes the experiments and results.
Finally, Section~\ref{conclusion} concludes the paper.

%% file: 2related_CA.tex
\section{Related Works}
\label{Review}

The rapid proliferation of generative models has exposed fundamental limitations of artifact-based deepfake detection, which remains reactive and struggles to generalize across novel architectures. To address these limitations, recent research has increasingly shifted toward proactive, source-centric watermarking strategies that embed verifiable provenance directly into generated content, enabling integrity verification and source attribution. Existing approaches span multiple paradigms, including model-specific fingerprints for attribution~\cite{yu2022artificialfingerprintinggenerativemodels}, semi-fragile marks for tamper detection~\cite{FaceSigns, yang2021faceguardproactivedeepfakedetection}, robust watermarking methods~\cite{HiDDeN, StegaStamp, MBRS} designed to withstand real-world distortions and GAN-based post-processing~\cite{ARWGAN}, hybrid fragile robust frameworks that support joint detection and tracing~\cite{SepMark}, and identity–semantic watermarking schemes for authenticity verification~\cite{wang2024robustidentityperceptualwatermark}. Collectively, these approaches demonstrate the potential of watermarking as a proactive mechanism for media authentication while avoiding disruptive perturbations and enabling provenance tracing.

Within this paradigm, several architectures address different challenges through specialized designs. \cite{Baseline} and FaceSigns \cite{FaceSigns} focus on deepfake detection through controlled fragility, embedding invisible bit strings that become non-recoverable when malicious facial manipulations occur~\cite{FaceSigns}. To mitigate watermark removal vulnerabilities, \cite{Baseline} introduce an adversarial network that improves resilience against both white-box and black-box attacks. SepMark \cite{SepMark} addresses the joint detection–attribution problem using a dual-channel architecture, enabling simultaneous robust source tracing and semi-robust tamper localization with a single embedded watermark. LampMark \cite{LampMark} instead focuses on improving generalization against hyper-realistic synthetic media by proposing a training-free, landmark-based perceptual watermark that exploits structural consistency changes introduced during deepfake generation, comparing recovered watermarks against dynamically generated structural identifiers for reliable detection. Despite these advances, existing approaches rarely provide transparent, human-interpretable forensic evidence and often lack principled multi-objective frameworks that jointly optimize imperceptibility, robustness, and source discriminability.

\textbf{Positioning of our work:} Despite significant progress, several challenges remain. Many frameworks require retraining or manual reconfiguration to handle emerging deepfake generators, limiting scalability, while multi-source watermarking remains difficult when embedding generator-specific identifiers without degrading visual fidelity. In addition, existing systems rarely provide dual-verification mechanisms that combine deepfake detection with reliable source tracing and clear, human-interpretable forensic evidence. Unlike conventional watermarking approaches that embed arbitrary payload bits, we introduce a source-conditioned watermark identity formulation in which the watermark itself encodes the generating media source and directly conditions the embedding process, enabling scalable multi-source provenance encoding within a unified framework. To address these limitations, we propose a scalable watermarking framework that embeds visual logos and machine-readable identifiers for different deepfake generators alongside markers for authentic content. The architecture employs a dual-mode forensic decoder that produces both human-verifiable watermark recovery and automated source classification, enabling robust attribution under real-world distortions and adversarial attacks while jointly optimizing imperceptibility, robustness, and source discriminability.

%% file: 3method_CA.tex
\begin{figure*}[!t]
    \centering
    \includegraphics[width=0.9\textwidth]{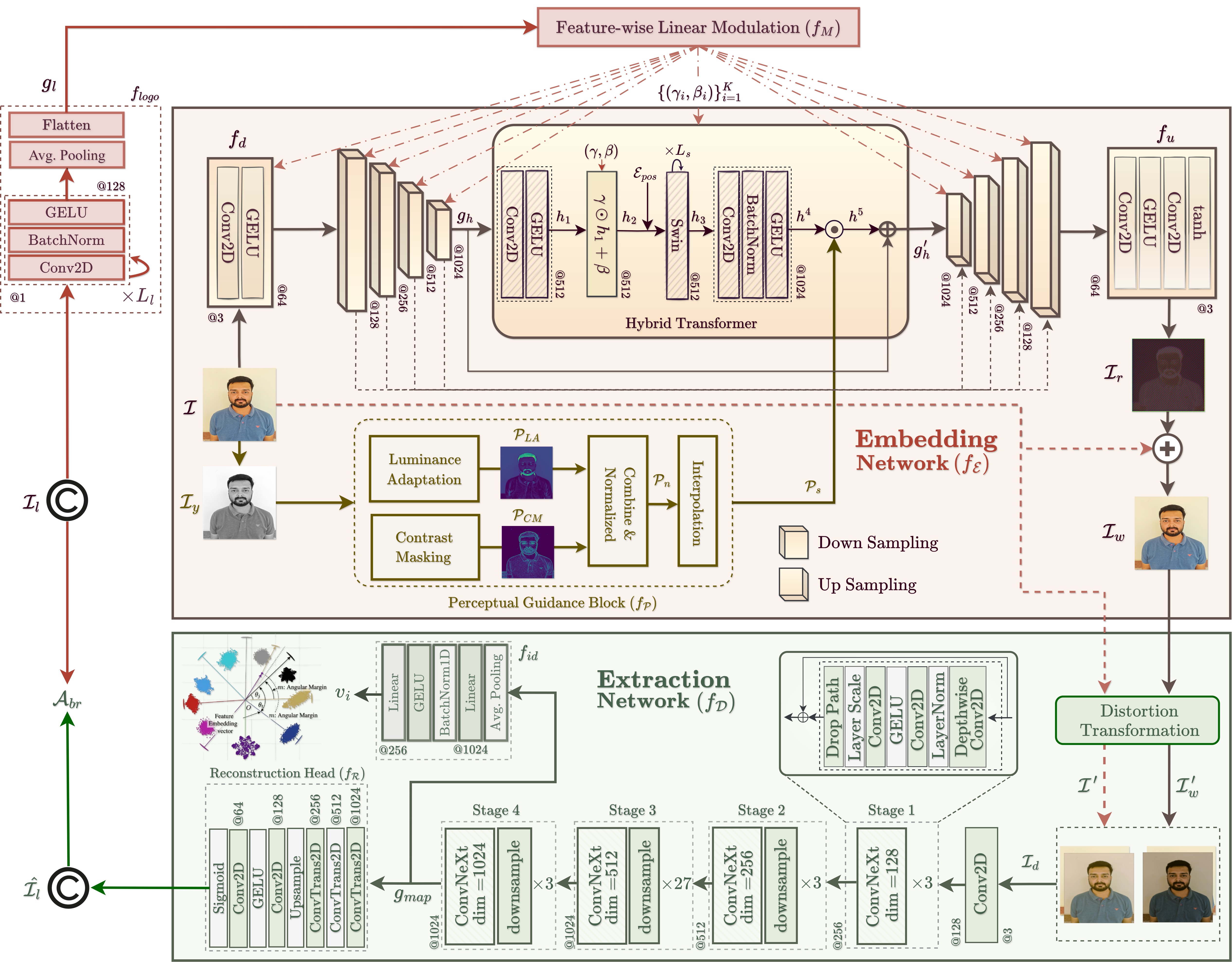}
    \caption{\small Workflow of the proposed framework, {\titleabbr} }
    \label{fig:encoder}
\end{figure*}

\section{Proposed Methodology}
\label{method}
We propose {\titleabbr}, a proactive deepfake defense framework that enables source-aware authentication by embedding verifiable watermarks at the time of content generation (Fig.~\ref{fig:encoder}). {\titleabbr} integrates feature-wise linear modulation with perceptual guidance to achieve both robustness and visual imperceptibility. It consists of two core components: a watermark embedding network ($f_{\mathcal{E}}$), which embeds source-specific watermarks into images, and a watermark extraction network ($f_\mathcal{D}$), which reliably recovers the embedded watermark under various distortions. 


\subsection{Overview and Problem Formulation}

Our framework operates on a collection of cover images $\mathcal{I} \in \mathbb{R}^{H \times W \times C}$, each originating from a distinct source and paired with unique source-specific watermark $\mathcal{I}_l \in \mathbb{R}^{h \times w \times c}$. The embedding network $f_{\mathcal{E}}$ integrates these watermarks into the corresponding images to generate visually indistinguishable outputs $\mathcal{I}_w$, while the extraction network $f_\mathcal{D}$ recovers the embedded watermark such that $\hat{\mathcal{I}}_{l} \approx \mathcal{I}_l$. This formulation enables scalable multi-source provenance encoding with high visual fidelity and reliable watermark recovery. Here, $(H, W, C) = (256, 256, 3)$ and $(h, w, c) = (64, 64, 1)$ denote the dimensions of the cover image and watermark, respectively.


Watermark embedding is formulated as an additive residual learning task, where the embedding network $f_{\mathcal{E}}$ generates a residual map $\mathcal{I}_r$ conditioned on both the cover image $\mathcal{I}$ and watermark $\mathcal{I}_l$. The watermarked image is obtained as
$\mathcal{I}_w = f_{clip}(\mathcal{I} + \mathcal{I}_r; 0, 1)$,
where $f_{clip}$ constrains pixel intensities to the normalized range $[0,1]$.

To ensure visual imperceptibility, a perceptual guidance map $\mathcal{P}_n$ is derived from $\mathcal{I}$ based on human visual priors. This map adaptively allocates stronger perturbations to textured or high-frequency regions while suppressing changes in perceptually sensitive areas, enabling effective concealment of the watermark.



The extraction network $f_\mathcal{D}$ operates on distorted inputs $\mathcal{I}_d$ to recover the embedded watermark:
$\hat{\mathcal{I}}_l = f_\mathcal{D}(\mathcal{I}_d)$.
The recovered watermark serves both for reconstruction and for source identification, enabling reliable attribution of the image origin.

\subsection{Watermark Embedding Network ($f_{\mathcal{E}}$)}
\label{subsec:embedding_nw}

The embedding network $f_{\mathcal{E}}$ adopts a U-Net backbone integrated with a Hybrid Transformer bottleneck, enabling it to capture both fine-grained local features and global contextual dependencies. In addition, Feature-wise linear modulation $f_\textit{M}$ enables joint conditioning on the cover image and the source-specific watermark identity while being guided by the perceptual guidance block $f_\mathcal{P}$.


\subsubsection{Source-Conditioned Watermark Encoding}
To achieve source-specific watermark embedding, the watermark $\mathcal{I}_l$ is first processed by a lightweight logo encoder $f_{\textit{logo}}$ composed of $L_l$ number of stacked $\mathrm{Conv2D}$ layers with batch normalization and GELU activations. The resulting feature maps are aggregated using adaptive average pooling followed by flattening to produce a compact global feature representation $g_\textit{l} \in \mathbb{R}^{128}$. Here, we fix $L_l=5$, empirically. 
Since watermark identities are represented through learned latent embeddings, the framework can scale to large numbers of media sources without redesigning the embedding architecture. 

%
Now $g_\textit{l}$ is passed for feature-wise linear modulation module $f_\textit{M}$ to predict parameters $\{({\gamma}_i,{\beta}_i)\}_{i=1}^{K}$ used to condition multiple layers of the watermark encoder $f_\mathcal{E}$. Here, $K=10$. 
Given a feature map $\mathbf{x}_i$ at the $i^{\text{th}}$-encoder stage, $f_\textit{M}$-conditioning is defined as: 
$f_\textit{M}(\mathbf{x}_i; {\gamma}_i, {\beta}_i)
= {\gamma}_i \odot \mathbf{x}_i + {\beta}_i$, 
where $\odot$ denotes element-wise multiplication, and ${\gamma}_i$ and ${\beta}_i$ represent the learned scaling and shifting parameters \cite{FiLM2018}. 
$f_\textit{M}$ conditioning provides lightweight channel-wise modulation of encoder features across multiple resolutions, enabling the network to learn watermark-specific transformations and adaptively modulate embedding patterns according to the source identity with minimal parameter overhead.
Appendix A and Fig. B.1 (\textit{supp. file}) analyze $f_M$-based conditioning, highlighting its role in learning source-specific style embeddings.



\subsubsection{Content-Adaptive Embedding via Perceptual Guidance} 
To balance imperceptibility and robustness, we engage a perceptual guidance block ($f_\mathcal{P}$) that estimates spatial perceptual tolerance from luminance structure. The perceptual guidance mechanism follows principles from human visual system models used in perceptual watermarking literature \cite{JND2017}. 
Given a cover image  $\mathcal{I} {\in} \mathbb{R}^{H {\times} W {\times} C}$, it is first converted to YUV color space, and the luminance channel $\mathcal{I}_y {\in} \mathbb{R}^{H {\times} W {\times} C_y}$ is extracted. Here, $C_y=1$. 
Luminance adaptation is computed by estimating the local background luminance $\mathbf{L}_{bg}$ from the luminance channel using a $5{\times}5$ smoothing kernel $\mathcal{K}_l$, i.e., $\mathbf{L}_{bg} = \mathcal{K}_l {*} \mathcal{I}_y$, where $*$ denotes convolution operation. 
The perceptual luminance adaptation component $\mathcal{P}_{LA}$ is then defined as:  
\begin{equation} \small 
    \mathcal{P}_{LA} = 
    \begin{cases}
p_1\left(1- \left({\mathbf{L}_{bg}}/{p_2} \right)^{{1}/{2}}\right)+p_3; & 0 \le \mathbf{L}_{bg} \le p_2 \\
{p_3}(\mathbf{L}_{bg}-p_2)/ ({p_2+1}) + p_3; & p_2 < \mathbf{L}_{bg} \le 255
\end{cases}
\end{equation} 
where $p_1$, $p_2$ are scaling factors, and $p_3$ acts a threshold. 
In parallel, contrast masking is obtained from spatial gradients computed using Sobel filters $\mathcal{O}_x$ and $\mathcal{O}_y$ to estimate local edge strength. 
Specifically, the gradient magnitude is computed as 
$\nabla_s {=} \left({(\mathcal{O}_x {*} \mathcal{I}_y)^2 {+} (\mathcal{O}_y {*} \mathcal{I}_y)^2}\right)^{1/2}$, 
which captures the spatial variation of luminance. 
Based on this gradient response, the perceptual contrast masking component $\mathcal{P}_{CM}$ is computed as: 
$\mathcal{P}_{CM} = ({p_4(\nabla_s)^{p_5}}) / ({(\nabla_s)^2 + ({p_6})^2})$,  
where $p_4$ is a scaling factor and $p_6$ is a stabilization constant controlling the saturation of masking strength. 
Empirically, we set 
$p_1 = 17$, 
$p_2 = 127$, 
$p_3 = 3$, 
$p_4 = 1.872$, 
$p_5 = 8$, and 
$p_6 = 26$. 
Now, $\mathcal{P}_{LA}$ and $\mathcal{P}_{CM}$ are combined to form a perceptual map: 
$\mathcal{P}=\mathcal{P}_{LA}+\mathcal{P}_{CM}-\lambda_{1}\min(\mathcal{P}_{LA},\mathcal{P}_{CM})$, 
where $\lambda_{1}$ is a tunable hyperparameter. 
%
The fused perceptual map $\mathcal{P}$ is rectified using a ReLU activation and 
normalized to produce the perceptual guidance map 
$\mathcal{P}_n {=} f_{clip}\left(\lambda_2 \mathrm{ReLU}(\mathcal{P});0,3\right)$, 
where $\lambda_2$ rescales the perceptual thresholds. 
Now, $\mathcal{P}_n\in\mathbb{R}^{H\times W \times 1}$ is injected into the hybrid transformer bottleneck to modulate feature representations based on local perceptual sensitivity.




\subsubsection{Hybrid Transformer Bottleneck} 

As illustrated in Fig.~\ref{fig:encoder}, the input image $\mathcal{I} {\in} \mathbb{R}^{H {\times} W {\times} C}$ is first processed by an initial feature extractor $f_\textit{d}$, implemented as $\mathrm{Conv2D}$ followed by a GELU activation, to extract low-level features. The resulting feature representation is then progressively encoded through a series of down-sampling blocks to obtain feature map $g_h {\in} \mathbb{R}^{16 {\times} 16 {\times} 1024 }$, which serves as the input to the hybrid transformer bottleneck to capture both local structural cues and long-range spatial dependencies. This feature map $g_h$ is then passed through $\mathrm{Conv2D}$, followed by GELU, which first projects the feature into the transformer embedding space, producing 
$h_1$. 
The representation is then modulated through a $f_\textit{M}$ layer using dynamically generated parameters $(\gamma,\beta)$: 
$h_2 {=} \gamma \odot h_1 + \beta$, 
where $\odot$ denotes element-wise scaling;
$h_1, h_2 \in  \mathbb{R}^{16 {\times} 16 {\times} 512 }$. 

The $f_\textit{M}$ conditioned features are flattened and enriched with learnable positional embeddings $\mathcal{E}_{pos}$ before being processed by a stack of $L_s$ number of Swin transformer layers \cite{liu2021swin}: 
$h_3 {=} \mathcal{S}^{(L_s)}(h_2 + \mathcal{E}_{pos})$, 
where $\mathcal{S}^{(L_s)}(\cdot)$ denotes the Swin transformation with $L_s$ layers; $ h_3  {\in} \mathbb{R}^{256 {\times} 512 } $. 
Here, we empirically set $L_s{=}4$. 
The resulting representation $h_3$ is reshaped and projected back to the original channel dimension using a $\mathrm{Conv2D}$ followed by batch norm and GELU, yielding $h_4$. 

Now the perceptual guidance map $\mathcal{P}_n$ is converted to $\mathcal{P}_s$ using bilinear interpolation that 
scales $h_4$ as:  
$h_5 = (1 {-} \lambda_3 {\cdot} \mathcal{P}_s).h_4$, 
where $\lambda_3$ regulates the trade-off between suppressing perturbations in perceptually fragile regions and amplifying embedding strength in visually tolerant areas; $h_4, h_5 \in \mathbb{R}^{16 {\times} 16 {\times} 1024 }$; $\mathcal{P}_s \in \mathbb{R}^{16 {\times} 16 {\times} 1}$.

To incorporate perceptual priors, $h_5$ is connected with the bottleneck input $g_h$ through a skip connection: 
$g'_h {=} g_h + h_5$; 
$g'_h \in \mathbb{R}^{16 {\times} 16 {\times} 1024}$. 

This perceptually guided bottleneck allocates stronger watermark embedding in textured or edge-rich regions while suppressing distortions in visually sensitive smooth areas. The decoder stage then mirrors the encoder structure with $f_\textit{M}$ conditioning at each resolution level to progressively reconstruct the residual watermark map $\mathcal{I}_r$. Specifically, the decoded feature map is passed through a refinement block $f_u$ consisting of a $3{\times}3$ $\mathrm{Conv2D}$ followed by GELU activation and a $1{\times}1$ $\mathrm{Conv2D}$ layer to project the features into the residual space. A $\tanh$ activation is then applied to constrain the residual values within a bounded range. The resulting residual map is further scaled by a learnable factor $\lambda_4$ to control the embedding strength, yielding the final residual watermark representation: $\mathcal{I}_r = \lambda_4 {\cdot} f(\mathcal{I}, \mathcal{I}_l)$. 
Finally, the scaled residual $\mathcal{I}_r$ is added to the input image $\mathcal{I}$ to produce the watermarked image $\mathcal{I}_w$; 
$\mathcal{I}_w = \mathcal{I}_r {+} \mathcal{I}$. 
The scalar hyperparameters are empirically set to $\lambda_1 {=} 0.3$, $\lambda_2 {=} \tfrac{1}{30}$, and $\lambda_3 {=} 0.5$, while $\lambda_4$ is learned during training to control the overall embedding strength.
Appendix B further analyzes the contributions of $f_M$, $f_\mathcal{P}$, and the Swin transformer $\mathcal{S}$ in enabling effective watermark conditioning.
%
%


\subsection{Watermark Extraction Network ($f_\mathcal{D}$)}
The extractor $f_\mathcal{D}$ is designed to robustly recover the embedded watermark and reconstruct it from distorted image $\mathcal{I}_d$. As shown in Fig.~\ref{fig:encoder}, the network employs a ConvNeXt backbone \cite{ConvNext}, selected for its powerful hierarchical feature representation.


\subsubsection{Distortion Transformation}
To simulate realistic degradations arising from post-processing operations, distribution pipelines, and potential adversarial manipulations, we apply a set of stochastic, non-differentiable transformations $\mathcal{T}(\cdot)$ (e.g., compression, resizing, blurring, and noise injection). 
Formally, the transformations are applied to both the original image $\mathcal{I}$ and the watermarked image $\mathcal{I}_w$, yielding $\mathcal{I}_w'=\mathcal{T}(\mathcal{I}_w)$ and $\mathcal{I}'=\mathcal{T}(\mathcal{I})$. 
The resulting Image $\mathcal{I}_d=(\mathcal{I}'\,|\,\mathcal{I}_w')$ is then used for feature encoding.

\subsubsection{Feature Encoding}
The distorted input image $\mathcal{I}_d$ is first processed by a Conv2D layer with a $7{\times}7$ kernel and stride $4$ to generate an initial low-level embedding. This representation is subsequently passed through four hierarchical ConvNeXt stages. Each stage progressively enlarges the receptive field while enhancing the semantic richness of the representation. Following the ConvNeXt design, each stage employs depthwise convolutions, layer normalization, GELU activations, and skip connections that enable efficient feature extraction and stable optimization~\cite{ConvNext}. After the final ConvNeXt stage (Stage~4), a compact feature map $g_{\textit{map}} \in \mathbb{R}^{8 \times 8 \times 1024}$ is obtained. 
This feature map encodes high-level semantic information and is fed into two parallel branches: 
a watermark reconstruction head ($f_{\mathcal{R}}$) and 
a source identification head ($f_{id}$). 

\paragraph{Reconstruction Head}
The reconstruction head $f_{\mathcal{R}}$ progressively upsamples $g_{\textit{map}}$ through a convolutional decoder composed of multiple ConvTranspose2D layers followed by an upsampling block. This hierarchical decoding process gradually restores spatial resolution while preserving watermark-related structures. A final sigmoid activation produces the reconstructed binary watermark $\hat{\mathcal{I}}_{l}$, i.e., $\hat{\mathcal{I}}_{l} = f_{\mathcal{R}} (g_{\textit{map}})$.


\paragraph{Source Identification Head} 
In parallel, the feature map $g_{\textit{map}}$ is transformed into a compact embedding for source identification. Specifically, global average pooling aggregates spatial information to produce a 1024-dimensional feature vector, which is then projected through a linear layer to obtain a latent embedding $v_i = f_{id}(g_{\textit{map}})$; $v_i {\in} \mathbb{R}^{256}$. To learn discriminative source-aware representations, the embedding is optimized using an angular margin-based identification loss $\mathcal{L}_{id}$. The classification is performed using cosine similarity between the embedding vector $v_i$ and the corresponding class weights. 
This objective encourages embeddings from the same source to form compact clusters while increasing angular separation between different source identities. In conjunction with watermark reconstruction, this identification objective provides complementary supervision: reconstruction ensures faithful recovery of the embedded watermark, while the classification head enables reliable provenance attribution even when the watermark is partially degraded.

\subsection{Training Loss}
The proposed {\titleabbr} is trained under a composite objective that jointly optimizes for three criteria: (i) imperceptibility of the embedded watermark, (ii) robustness against distortions and attacks, and (iii) discriminative identity preservation. The total loss is formulated as:
\begin{equation}
\mathcal{L}_{\textit{total}} = w_{\textit{imp}} \mathcal{L}_{\textit{imp}} + w_{\textit{rob}} \mathcal{L}_{\textit{rob}} + w_{\textit{id}} \mathcal{L}_{\textit{id}}
\end{equation}
where $w_{\textit{imp}}$, $w_{\textit{rob}}$, and $w_{\textit{id}}$ are non-negative weights balancing the respective objectives; 
$ w_{\textit{imp}} + w_{\textit{rob}} + w_{\textit{id}} =1$.


\subsubsection{Imperceptibility Loss ($\mathcal{L}_{\textit{imp}}$)}
To preserve perceptual fidelity between the cover image $\mathcal{I}$ and the watermarked image $\mathcal{I}_w$, we use a weighted combination of pixel-wise $L_1$ loss and the Learned Perceptual Image Patch Similarity (LPIPS) metric:
\begin{equation}
\mathcal{L}_{\textit{imp}} = w_1 \, \mathcal{L}_{L_1}(\mathcal{I}, \mathcal{I}_w) + w_{\textit{lpips}} \, \mathcal{L}_{\textit{lpips}}(\mathcal{I}, \mathcal{I}_w),
\end{equation}
where $\mathcal{L}_{\textit{lpips}}$ captures perceptual similarity more consistently with human vision, complementing the low-level fidelity enforced by $L_1$ loss; $w_1 + w_{\textit{lpips}} = 1$. 

\subsubsection{Robustness Loss ($\mathcal{L}_{\textit{rob}}$)}
The robustness objective ensures accurate recovery of the watermark $\mathcal{I}_l$ from $\mathcal{I}_d$.  
Let $\hat{\mathcal{I}}_l = f_\mathcal{D}(\mathcal{I}_d)$ denote the reconstructed watermark logo. 
The robustness loss is defined as:
\begin{equation}
\scalebox{0.80}{$
\mathcal{L}_{\textit{rob}} = \mathbb{E}_{\mathcal{A}} \left[ 
w_2 \, \mathcal{L}_{L_1}(\hat{\mathcal{I}}_l, \mathcal{I}_l) +
w_{\textit{ssim}} \, \mathcal{L}_{\textit{ssim}}(\hat{\mathcal{I}}_l, \mathcal{I}_l) +
w'_{\textit{lpips}} \, \mathcal{L}_{\textit{lpips}}(\hat{\mathcal{I}}_l, \mathcal{I}_l)
\right]
$}
\end{equation} 
where $\mathcal{L}_{\textit{ssim}}$ promotes structural similarity, $\mathbb{E}_{\mathcal{A}}[\cdot]$ denotes expectation over sampled attack conditions; 
$w_2 + w_{\textit{ssim}} + w'_{\textit{lpips}} = 1$. 

\subsubsection{Identification Loss ($\mathcal{L}_{\textit{id}}$)}
We employ the additive angular margin loss \cite{Arcface} to ensure intra-class compactness and inter-class separability of identity vectors. 
On the hypersphere, $\mathcal{L}_{\textit{id}}$ imposes an angular margin $m$ between classes. For a batch of $L_2$-normalized vectors $\{v_i\}_{i=1}^N$ with class labels $\{y_i\}_{i=1}^C$, the loss is given by:
\begin{equation}
\scriptsize
\mathcal{L}_{\textit{id}} = \frac{-1}{N} \sum_{i=1}^N  
\log \frac{\exp\left(\lambda_5 {\cdot} \cos(\theta_{y_i} + m)\right)}
{\exp\left(\lambda_5 {\cdot} \cos(\theta_{y_i} + m)\right) + 
\sum_{\substack{j=1 \\ j\neq y_i}}^{C} \exp\left(\lambda_5 {\cdot} \cos(\theta_j)\right)}
\end{equation}
where $\theta_j$ is the angle between $v_i$ and the class weight, $\lambda_5$ is a scale parameter, $m$ is the additive angular margin, and $C$ is the total number of watermark classes plus 1 (for ``\textit{no watermark}''). This formulation encourages intra-class compactness and inter-class separation on the hypersphere. 
Empirically, we fix $m=0.4$, mini-batch size $N=16$, $\lambda_5 = 30$. 
To train the overall {\titleabbr} framework, we engaged AdamW optimizer.

%% file: 5result_CA.tex
\section{Experiments and Discussions}
\label{Exp}

This section presents the experimental evaluation of {\titleabbr}. All experiments were implemented in PyTorch 2.2.2 with CUDA 11.8 and conducted on a system equipped with an Intel Xeon W7-2495X (24 cores), 256 GB RAM, and an NVIDIA RTX A6000 GPU with 48 GB VRAM.

\subsection{Datasets} \label{subsec:dataset}
To evaluate the proposed framework under diverse manipulation scenarios, we conducted experiments on multiple publicly available datasets: IndicSideFace \cite{IndicSideFace2025}, FaceForensics (FF) \cite{rossler2018faceforensics}, FaceForensics++ (FF++) \cite{rossler2019faceforensics++}, and CelebA \cite{liu2015faceattributes}.
The IndicSideFace dataset \cite{IndicSideFace2025} served as our primary dataset and contains 984 genuine images from 164 Indian subjects across frontal and side-profile views, along with 21648 deepfake images generated using five identity-swapping methods (Ghost \cite{ghost2022}, SimSwap \cite{simswap2020}, SimSwap++ \cite{simswap++2023}, FaceDancer \cite{facedancer2023}, and InsightFace \cite{InsightFace}) and one attribute-modification tool (FaceApp \cite{faceapp}). To assess cross-dataset generalization, we further evaluated {\titleabbr} on FF and FF++, sampling 60 frames per video with an 80:20 train-test split. Additionally, the CelebA \cite{liu2015faceattributes} dataset was used to generate additional watermarked facial samples by sampling 202599 images across more than 200 identities with a 70:30 train-test split. 
For logo-conditioned watermarking, LOGO-30K \cite{logo30k} was used, where each logo acts as a unique identifier representing a specific manipulation source or platform, enabling structured watermark embedding and systematic source attribution. This multi-dataset evaluation protocol allows the framework to be evaluated under both controlled and cross-domain conditions, enabling comprehensive analysis of watermark robustness and source attribution capability.

\subsection{Evaluation Criteria}
We evaluate {\titleabbr} according to three key criteria: 
\textit{imperceptibility}, which measures the visual fidelity of the watermarked image; 
\textit{robustness}, which evaluates the ability to recover the embedded watermark after the host image undergoes distortions; and 
\textit{source identification}, which assesses the model’s capability to correctly determine the origin of an image.

\subsubsection{Imperceptibility} 
This quantifies the visual distortion introduced during watermark embedding \cite{zhao2016loss}. 
We compare the original cover image $\mathcal{I}$ and the corresponding watermarked image $\mathcal{I}_w$ using three widely adopted perceptual metrics.

\paragraph{PSNR (Peak Signal-to-Noise Ratio)}
PSNR measures the ratio between the maximum possible pixel value ($\textit{max}_{\mathcal{I}}$) and the reconstruction distortion. 
For images normalized to $[0,1]$, $\textit{max}_{\mathcal{I}} = 1$.  
\begin{equation} \small
\text{PSNR}(\mathcal{I}, \mathcal{I}_w) 
= 10 \log_{10} \left({\textit{max}_{\mathcal{I}}^2}/{\text{MSE}}\right)
\end{equation} 
where, $\text{MSE}
=
\frac{1}{H \cdot W \cdot C}
\sum_{i=1}^{H}
\sum_{j=1}^{W}
\sum_{k=1}^{C}
(\mathcal{I}^{\langle i,j,k\rangle}-\mathcal{I}_w^{\langle i,j,k\rangle})^2$; 
$\mathcal{I}, \mathcal{I}_w {\in} \mathbb{R}^{H {\times} W {\times} C}$; $i,j,k$ denote height, width, channel indices. 
Higher PSNR values indicate better visual fidelity.




\paragraph{SSIM (Structural Similarity Index Measure)} 
SSIM evaluates perceptual similarity by comparing luminance, contrast, and structural information across image patches between $\mathcal{I}$ and $\mathcal{I}_w$. 
\begin{equation} \scriptsize
\label{ssim} 
\resizebox{0.38\textwidth}{!}{$ 
\text{SSIM}(\mathcal{I}, \mathcal{I}_w) =  
\frac{(2\mu_{\mathcal{I}}\mu_{\mathcal{I}_w} + c_1) (2\sigma_{\mathcal{I},\mathcal{I}_w} + c_2)} 
{(\mu_{\mathcal{I}}^2 + \mu_{\mathcal{I}_w}^2 + c_1)(\sigma_{\mathcal{I}}^2 + \sigma_{\mathcal{I}_w}^2 + c_2)} 
$} 
\end{equation} 
where
$\mu_{\mathcal{I}}$, $\mu_{\mathcal{I}_w}$ denote pixel sample means of $\mathcal{I}$, $\mathcal{I}_w$, respectively; 
$\sigma_{\mathcal{I}}^2$, $\sigma_{\mathcal{I}_w}^2$ denote sample variances of $\mathcal{I}$, $\mathcal{I}_w$, respectively; 
$\sigma_{\mathcal{I},\mathcal{I}_w}$ represents their their sample co-variance; 
and $c_1$, $c_2$ are  stabilization constants. 
The final score is obtained by averaging SSIM across all image patches.

\paragraph{LPIPS (Learned Perceptual Image Patch Similarity)}
LPIPS measures perceptual similarity using deep features extracted from a pretrained network $\mathcal{N}$ \cite{lpips}. 
Feature maps from multiple layers are channel-wise normalized, and the weighted squared $L_2$ distance between corresponding activations is computed across spatial locations and channels. 
The final perceptual distance is obtained by averaging over spatial dimensions and aggregating across layers:

\begin{equation} \scriptsize
\text{LPIPS}(\mathcal{I},\mathcal{I}_w)
=
\sum_l
\frac{1}{H_l W_l}
\sum_{i=1}^{H_l}
\sum_{j=1}^{W_l}
\sum_{k=1}^{C_l}
w_l(k)
\|
\hat{\mathcal{N}}_l(\mathcal{I}^{\langle i,j,k\rangle})
-
\hat{\mathcal{N}}_l(\mathcal{I}_w^{\langle i,j,k\rangle})
\|_2^2
\end{equation}
where $\hat{\mathcal{N}}_l(\cdot)$ denotes normalized features at layer $l$; $H_l,W_l,C_l$ are the feature dimensions; and $w_l(k)$ are learned channel-wise weights. Lower LPIPS indicates higher perceptual similarity.

\subsubsection{Robustness} This measures the ability of the extractor to faithfully recover the embedded watermark logo $\mathcal{I}_l \in \mathbb{R}^{h \times w \times c}$ from a distorted watermarked image $\mathcal{I}'_w$. The quality of the recovered logo $\hat{\mathcal{I}}_{l}$ is evaluated using:

\paragraph{SSIM ($\mathcal{A}_{ssim}$)} 
SSIM (Eq.~\ref{ssim}) is applied to evaluate the structural integrity of the recovered logo $\hat{\mathcal{I}}_{l}$ relative to the original logo $\mathcal{I}_l$; 
$\mathcal{A}_{ssim} = \text{SSIM}(\mathcal{I}_l, \hat{\mathcal{I}}_{l}) $.

\paragraph{Bit Recovery Accuracy ($\mathcal{A}_{br}$)}
To quantitatively assess the fidelity of the recovered watermark signal, we employ $\mathcal{A}_{br}$ that measures the proportion of correctly reconstructed bits in the recovered watermark logo after binarization. 
\begin{equation}
\scriptsize 
\mathcal{A}_{br}({\mathcal{I}}_{l}, \hat{\mathcal{I}}_{l}) =
\frac{1}{h {\cdot} w}
\sum_{i=1}^{h} \sum_{j=1}^{w}
\left[ 1 - \left| B(\mathcal{I}_l^{\langle i,j \rangle}, \tau) - B(\hat{\mathcal{I}}_l^{\langle i,j \rangle}, \tau) \right| \right]
\end{equation}
where $B(\mathbf{x}^{\langle i,j\rangle},\tau)={\mathds{1}} \left[\mathbf{x}^{\langle i,j\rangle}>\tau\right]$ denotes a binarization operator that thresholds an image $\mathbf{x}$ at a scalar threshold $\tau$; 
${\mathds{1}}[.]$ is the indicator function. 
$\mathcal{A}_{br}$ measures the normalized agreement between binarized ground-truth and reconstructed logos, estimating bit-level recovery fidelity. $\mathcal{A}_{br}=1$ indicates perfect reconstruction and is appropriate for binary payloads requiring exact bit recovery.

\subsubsection{Source Identification}
This component evaluates the model’s ability to identify the generative or manipulation source of an input image using identification accuracy ($\mathcal{A}_{id}$). 
In the multi-class setting, predictions are summarized using a confusion matrix, where each entry $c_{ij}$ denotes the number of samples with ground-truth label $i$ predicted as class $j$. 
$\mathcal{A}_{id}$ is defined as:  
$\mathcal{A}_{id} = ({\sum_{i} c_{ii}})/({\sum_{i,j} c_{ij}})$, 
which represents the proportion of correctly classified samples. 
This metric reflects the model’s effectiveness in attributing images to their true generative source under different attack scenarios.



\input{Table/table1}

\input{Table/table_2}

\subsection{Quantitative Analysis}
\label{Quantitative Analysis} 
As shown in Table~\ref{tab:watermark_comparison}, {\titleabbr} achieves a superior trade-off between imperceptibility and robustness compared to existing watermarking methods, including DwtDct, DwtDctSvd \cite{DwtDctSvd}, MBRS \cite{MBRS}, StegaStamp \cite{StegaStamp}, RivaGAN \cite{RivaGAN}, SepMark \cite{SepMark}, and WAM \cite{JNDWAM}. 
In terms of imperceptibility, {\titleabbr} consistently produces high-fidelity watermarked images across all generators, achieving PSNR values above 50 dB, SSIM close to 1, and extremely low LPIPS scores. In contrast, existing methods operate at substantially lower PSNR ranges and exhibit higher perceptual distortion, indicating that the proposed perceptual embedding strategy effectively minimizes visible artifacts while preserving structural integrity. 
For pre-attack evaluation, {\titleabbr} achieves near-perfect bit recovery accuracy ($\mathcal{A}_{br}$) across all generators, matching the strongest baselines such as MBRS, StegaStamp, and WAM. Importantly, this performance is achieved while maintaining significantly higher visual fidelity, highlighting a clear advantage in the fidelity-robustness trade-off. 
Under post-attack conditions, {\titleabbr} maintains consistently high recovery performance across diverse distortions, including brightness, contrast, JPEG compression, Gaussian noise, and blur, with $\mathcal{A}_{br}$ remaining above 0.97 in nearly all cases. In comparison, several baselines exhibit noticeable degradation under challenging perturbations; for instance, StegaStamp shows reduced robustness under JPEG compression, while MBRS and SepMark experience performance drops under stochastic noise. Although WAM demonstrates strong robustness, {\titleabbr} achieves comparable recovery performance while significantly improving perceptual quality. 
Notably, the consistent robustness across heterogeneous distortions indicates that the model learns distortion-invariant watermark representations rather than overfitting to specific perturbation patterns. This property is critical for reliable deployment in real-world scenarios, where distortions are often unpredictable and compositional. 
These results demonstrate that {\titleabbr} effectively integrates perceptual-aware embedding with robust decoding, achieving high visual fidelity (PSNR ${>}51$ dB, SSIM ${>}0.998$, LPIPS ${<}0.001$) alongside reliable watermark recovery ($\mathcal{A}_{br} \geq 0.97$) across diverse generators and attack scenarios. This balance is essential for scalable and dependable source attribution in deepfake media.

\subsubsection{Performance of Proactive Detector on Cross Datasets} 
{\titleabbr} was evaluated on three widely used public benchmarks, namely FF \cite{rossler2018faceforensics}, FF++ \cite{rossler2019faceforensics++}, and CelebA \cite{liu2015faceattributes}, to assess both imperceptibility and robustness under cross-dataset settings, while trained on IndicSideFace \cite{IndicSideFace2025}. As summarized in Table~\ref{Public}, {\titleabbr} achieved consistently high perceptual fidelity, with PSNR values above 55~dB, SSIM scores exceeding 0.99, and LPIPS ranging from 0.0007 to 0.0019, indicating negligible visual distortion across diverse datasets. 
In addition to high imperceptibility, {\titleabbr} demonstrated strong robustness under various real-world distortions. Under Instagram-style photometric perturbations ($\mathbb{A}$: Aden, $\mathbb{B}$: Brooklyn, and $\mathbb{C}$: Clarendon), both structural similarity ($\mathcal{A}_{ssim}$) and bit recovery accuracy ($\mathcal{A}_{br}$) remained consistently at 0.99 across all filters and their combinations. Under JPEG compression, {\titleabbr} maintained $\mathcal{A}_{br}$ values of 0.96-0.98 at $50\%$ QF (quality factor), and near-perfect recovery at $75\%$ QF, demonstrating resilience even under strong compression. Similarly, under Gaussian blur with varying kernel sizes, both $\mathcal{A}_{ssim}$ and $\mathcal{A}_{br}$ remained stable at 0.99, indicating robustness to convolutional degradations.

These results comprehend that {\titleabbr} generalizes effectively across heterogeneous datasets while preserving high perceptual quality and robust watermark recovery. The consistent performance under diverse distortion scenarios highlights its suitability for practical and scalable deepfake forensics and proactive media integrity verification.

\begin{table*}[!t]
\centering
\caption{Comparison of watermark imperceptibility and robustness under Instagram-style photometric distortions on CelebA \cite{liu2015faceattributes}}
\label{instagram_filters} 
\begin{adjustbox}{width=0.75\textwidth}
\begin{tabular}{l|c|c| c|ccccccc|l|l} 
\hline 
\multirow{2}{*}{\textbf{Model}} & \multicolumn{2}{c|}{\textbf{Imperceptibility}} &  \textbf{Pre-Attack} & \multicolumn{9}{c}{\textbf{Post-Attack ($\mathcal{A}_{br} \uparrow$)}} \\ 
\cline{2-13}
\multicolumn{1}{c|}{} & \textbf{PSNR}$\uparrow$ & \textbf{SSIM}$\uparrow$ &  ($\mathcal{A}_{br} \uparrow$) & \textbf{$\mathbb{A}$} & \textbf{$\mathbb{B}$} & \textbf{$\mathbb{C}$} & \textbf{$\mathbb{A}{+}\mathbb{B}$} & \text{$\mathbb{B}{+}\mathbb{C}$} & \textbf{$\mathbb{A}{+}\mathbb{C}$} & \textbf{$\mathbb{A}{+}\mathbb{B}{+}\mathbb{C}$} & \textbf{JPEG} & \textbf{G-Blur} \\ 
\hline \hline
SemiFragile-DCT \cite{1286417} & 20.29 & 0.846 & 0.9943 & 0.9347 & 0.9579 & 0.9512 & 0.9239 & 0.9342 & 0.9474 & 0.9156 & 0.5386 & 0.7624 \\ \hline 
HiDDeN \cite{HiDDeN} & 24.96 & 0.928 & 0.9765 & 0.9461 & 0.9382 & 0.9413 & 0.9137 & 0.9129 & 0.9223 & 0.8979 & 0.6719 & 0.8497 \\ \hline
StegaStamp \cite{StegaStamp} & 28.64 & 0.922 & 0.9962 & 0.9948 & 0.9926 & 0.9909 & 0.9718 & 0.9628 & 0.9579 & 0.9413 & 0.9524 & 0.9914 \\ \hline
FaceSigns \cite{FaceSigns} & 36.08 & 0.975 & 0.9949 & 0.9945 & 0.9922 & 0.9915 & 0.9853 & 0.9786 & 0.9751 & 0.9636 & 0.9375 & 0.9824 \\ \hline
Nadimpalli et al. 
\cite{Baseline} & 35.57 & 0.970 & 0.9945 & 0.9939 & 0.9925 & 0.9919 & 0.9887 & 0.9854 & 0.9869 & 0.9718 & 0.9452 & 0.9918 \\ \hline
{\titleabbr} & \textbf{55.89} & \textbf{0.999} & \textbf{0.9974} & \textbf{0.9970} & \textbf{0.9967} & \textbf{0.9963} & \textbf{0.9932} & \textbf{0.9898} & \textbf{0.9942} & \textbf{0.9967} & \textbf{0.9955} & \textbf{0.9968} \\ \hline 

\multicolumn{13}{r}{Instagram filters: $\mathbb{A}$ (Aden), $\mathbb{B}$ (Brooklyn), $\mathbb{C}$ (Clarendon)
}
\end{tabular}
\end{adjustbox}
\end{table*}

\subsubsection{Robustness under Photometric Filter Distortions} 
We evaluated the robustness of {\titleabbr} under real-world photometric distortions by applying Instagram-style filters ($\mathbb{A}$: Aden, $\mathbb{B}$: Brooklyn, and $\mathbb{C}$: Clarendon, and their combinations) to images from the CelebA dataset \cite{liu2015faceattributes}. 
As shown in Table~\ref{instagram_filters}, {\titleabbr} obtained substantially higher imperceptibility compared to existing baselines, with the highest PSNR (55.89 dB) and SSIM (0.999), indicating minimal visual distortion after watermark embedding. 
More importantly, {\titleabbr} demonstrated consistently strong robustness under both individual and compositional filter perturbations. 
While prior methods exhibit noticeable degradation in $\mathcal{A}_{br}$ from pre-attack to $\mathbb{A}{+}\mathbb{B}{+}\mathbb{C}$ post-attack 
(e.g., SemiFragile-DCT \cite{1286417} drops to 0.9156, 
HiDDeN \cite{HiDDeN} to 0.8979, and 
FaceSigns \cite{FaceSigns} and \cite{Baseline} to 0.9636 and 0.9718, respectively), {\titleabbr} maintains near-perfect recovery, with $\mathcal{A}_{br}$ consistently above 0.99 across all filter settings. This includes challenging scenarios involving multiple filter compositions, where nonlinear photometric transformations typically degrade watermark integrity. 
Furthermore, {\titleabbr} exhibited strong resilience under additional distortions such as JPEG compression and Gaussian blur, achieving recovery accuracies of 0.9955 and 0.9968, respectively, on CelebA \cite{liu2015faceattributes}. 

These results highlight the robustness of the learned embedding and decoding strategy in preserving watermark information under diverse real-world transformations. Notably, the strong performance under filter compositions suggests that {\titleabbr} learns invariant watermark representations rather than overfitting to specific distortion types, thereby effectively balancing high imperceptibility with robust watermark recovery and enabling strong generalization under complex photometric distortions commonly encountered in social media pipelines.

\input{Image/Fig_2_Qualitative_Analysis/Q_Analysis}

\subsection{Qualitative Analysis}
\label{Qualitative Analysis}

{\titleabbr} embeds a source-conditioned watermark $\mathcal{I}_l$ into images, enabling fine-grained provenance encoding and reliable generator-level attribution. Each image, whether genuine or manipulated, is associated with a distinct binary watermark that serves as a persistent and verifiable source signature. 
Fig.~\ref{Q1} presents qualitative results demonstrating the end-to-end embedding and recovery pipeline on \cite{IndicSideFace2025}. The input image $\mathcal{I}$ and corresponding watermark $\mathcal{I}_l$ are jointly processed, while the perceptual map $\mathcal{P}_n$ identifies visually tolerant regions that guide imperceptible embedding. The bottleneck representation $g_h'$ captures the fused content-watermark features, and the residual map $\mathcal{I}_r$ reveals that embedding perturbations are spatially localized and low in magnitude. The resulting watermarked image $\mathcal{I}_w$ remains visually indistinguishable from the original, even after undergoing distortions to produce $\mathcal{I}_d$. Crucially, the recovered watermark $\hat{\mathcal{I}}_l$ retains strong structural fidelity to the original logo.

Across diverse poses, textures, and imaging conditions, the model consistently concentrates watermark energy in perceptually insensitive regions, ensuring minimal visual degradation while preserving robustness. The stability of reconstruction under distortions demonstrates that the learned representation captures invariant watermark features rather than distortion-specific artifacts. This unified embedding-decoding behavior enables reliable source attribution and interpretable forensic validation, highlighting the practical viability of {\titleabbr} for real-world media authentication.

\begin{figure*}[!ht]
    \centering
    \includegraphics[width=0.7\textwidth]{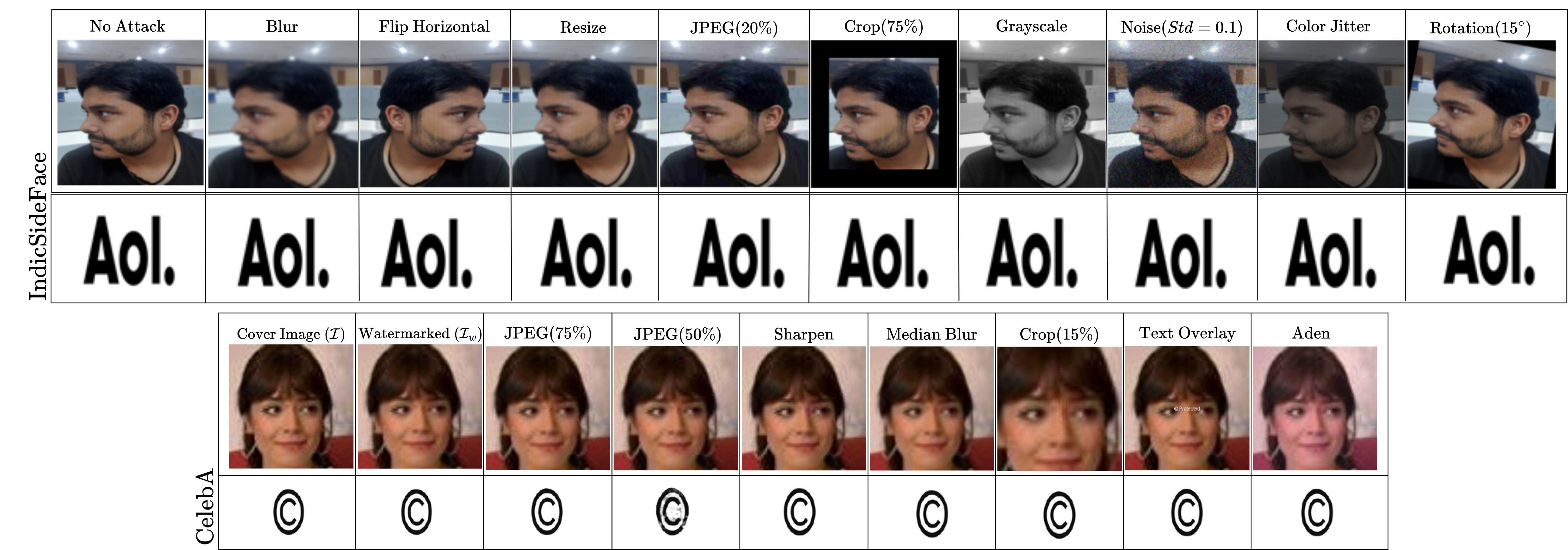}
\caption{\small Robustness evaluation of {\titleabbr} under diverse real-world distortions. 
\textit{Top:} The embedded ``\textbf{Aol.}” logo is perfectly recovered 
from an IndicSideFace \cite{IndicSideFace2025} sample after multiple distortions. 
\textit{Bottom:} The ``\textbf{\copyright}” watermark is recovered from CelebA \cite{liu2015faceattributes} sample after distortions.}
    
    \label{Attack}
\end{figure*}



\begin{figure}[!t]
    \centering
    \begin{tabular}{c|c}
       {\tiny(a)}\includegraphics[width=0.22\textwidth]  {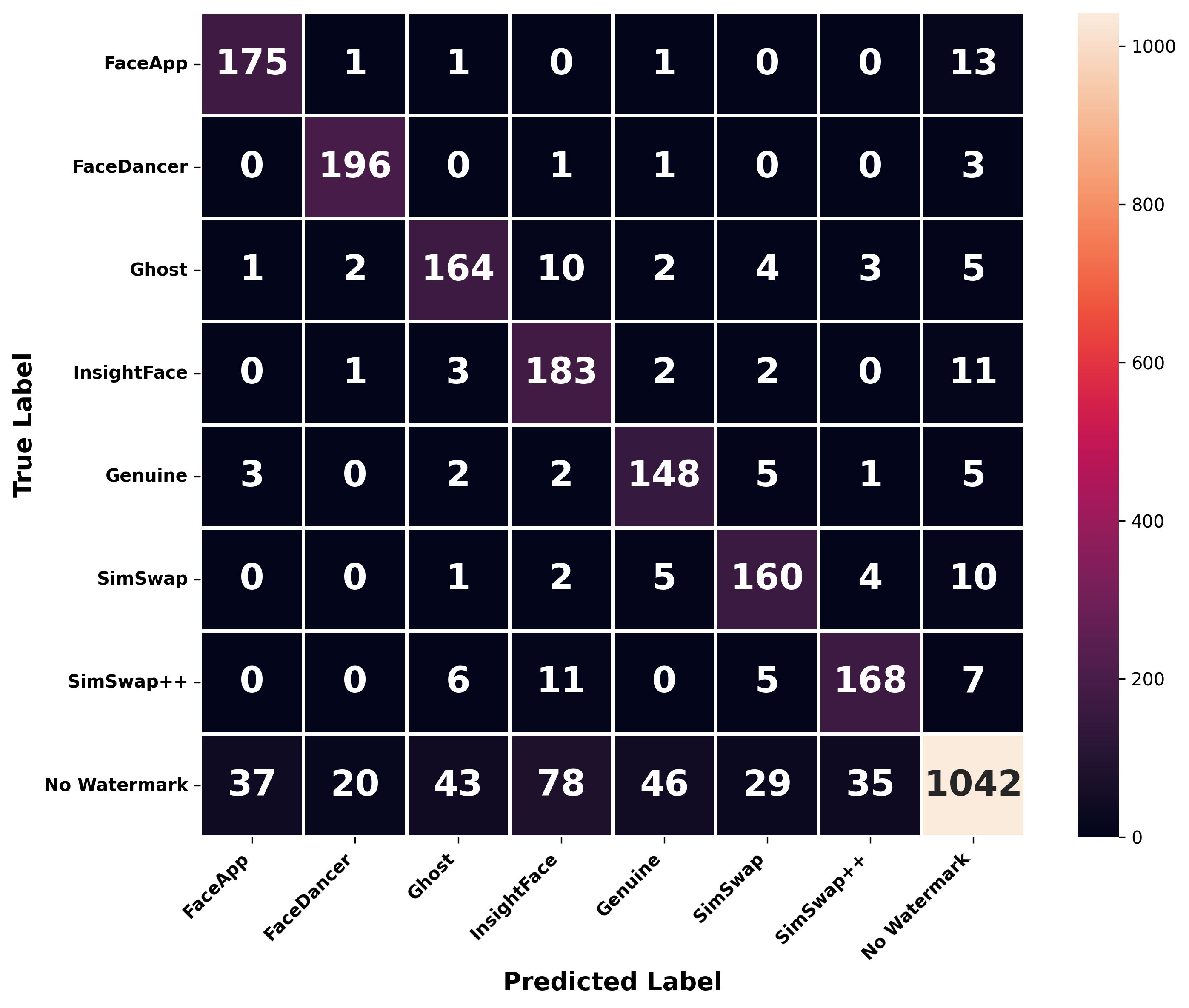} &
       {\tiny(b)}\includegraphics[width=0.22\textwidth]{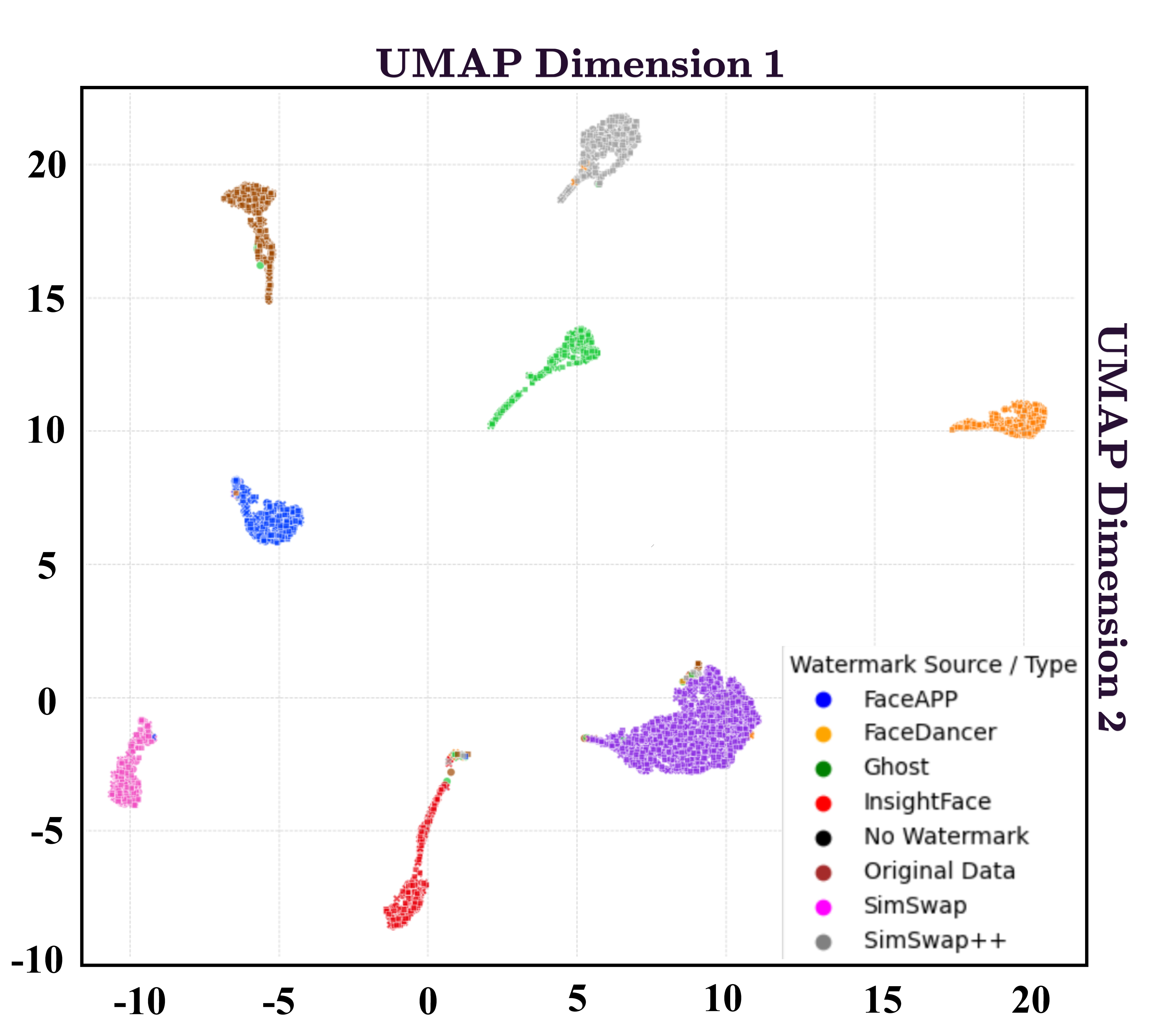}  \\ 
    \end{tabular}
    \caption{\small (a) Confusion matrix for source identification, (b) UMAP of extractor features between deepfake generators and originals.}
\label{Confusion_Matrix}
\end{figure}



%


\subsubsection{Robustness under Diverse Real-World Distortions} 
We comprehensively evaluate the robustness of {\titleabbr} under a wide spectrum of real-world distortions, with qualitative results illustrated in Fig.~\ref{Attack}. The evaluation encompasses geometric transformations (rotation, cropping, flipping, resizing), photometric and color variations (noise, color jitter, grayscale), filtering operations (blurring and sharpening), and severe lossy compression. As observed, the embedded watermark is consistently recovered with high fidelity across all scenarios on both IndicSideFace \cite{IndicSideFace2025} and CelebA \cite{liu2015faceattributes}, achieving $\mathcal{A}_{ssim}$ of approximately 0.99. Notably, the method remains stable even under challenging conditions such as aggressive JPEG compression, substantial cropping, and complex perturbations, including text overlays and Instagram-style filters. These results demonstrate that the learned embedding-decoding mechanism captures distortion-invariant watermark representations, ensuring reliable recovery under diverse transformations. Overall, the strong resilience across heterogeneous distortions highlights the practical suitability of the proposed framework for real-world media authentication and content protection.

\subsubsection{Watermark-Aware Source Attribution and Feature Space Analysis} 
To assess the discriminative capability of the learned representations, we extend the evaluation to a multi-class source attribution task, where each image is assigned to its exact generative origin. The label space includes multiple manipulation methods (e.g., FaceApp, FaceDancer, Ghost, InsightFace, SimSwap, and SimSwap++), along with two control categories: watermarked genuine, and no-watermark class. 
The performance of {\titleabbr} is presented using a confusion matrix and latent space visualization in Fig.~\ref{Confusion_Matrix}:(a)-(b). The confusion matrix exhibits strong diagonal dominance with minimal cross-class confusion, indicating reliable separation among generators, with an overall identification accuracy $\mathcal{A}_{id}$ of 84.06\%. Notably, the no-watermark class is distinctly recognized, reducing false positives and ensuring high specificity, which is critical for real-world forensic deployment. 
The UMAP projection in Fig.~\ref{Confusion_Matrix}:(b) reveals a well-structured feature space, where samples form compact and well-separated clusters across sources. This indicates that {\titleabbr} captures generator-specific signatures beyond watermark presence, enabling discriminative and robust source attribution under challenging conditions.

Additional analyses on semantic-style disentanglement, margin-based losses, and texture effects on $\mathcal{I}_r$ are presented in Appendices C-E (\textit{supp. file}).


%% file: Table/table1.tex
\begin{table*}[!t]
\centering
\footnotesize

\caption{Quantitative comparison of watermarking methods across multiple deepfake generators on IndicSideFace \cite{IndicSideFace2025}}
\label{tab:watermark_comparison} 
\begin{adjustbox}{width=0.75\textwidth}
\begin{tabular}{c|l|ccc |  c|ccccc}

\hline
\multirow{2}{*}{\textbf{Generator}} & \multirow{2}{*}{\textbf{Methods}} & \multicolumn{3}{c|}{\textbf{Imperceptibility}} & \textbf{Pre-Attack}  & \multicolumn{5}{c}{\textbf{Post-Attack ($\mathcal{A}_{br} \uparrow$)}} \\ 
\cline{3-5} 
\cline{7-11}
& & \multicolumn{1}{l}{\textbf{PSNR $\uparrow$}} & \multicolumn{1}{l}{\textbf{SSIM $\uparrow$}} & \textbf{LPIPS $\downarrow$} & ($\mathcal{A}_{br} \uparrow$) & \multicolumn{1}{l}{\textbf{Brightness}} & \multicolumn{1}{l}{\textbf{Contrast}} & \multicolumn{1}{l}{\textbf{JPEG}} & \multicolumn{1}{l}{\textbf{G-Noise}} & \textbf{G-Blur} \\ 
\hline \hline 
\multirow{7}{*}{\rotatebox{90}{{FaceApp} \cite{faceapp}}} & DwtDct & 44.40 & 0.981 & 0.0014 & 0.75 & 0.74 & 0.74 & 0.74 & 0.74 & 0.75 \\ 
 & DwtDctSvd \cite{DwtDctSvd} & 40.92 & 0.981 & 0.0048 & 0.87 & 0.74 & 0.74 & 0.75 & 0.82 & 0.82 \\ 
 & MBRS \cite{MBRS} & 43.72 & 0.989 & 0.0024 & \textbf{0.99} & 0.83 & 0.91 & 0.88 & 0.87 & 0.98 \\
 & StegaStamp \cite{StegaStamp} & 34.97 & 0.966 & 0.0254 & \textbf{0.99} & \textbf{0.99} & \textbf{0.99} & 0.80 & 0.95 & \textbf{0.99} \\ 
 & RivaGAN \cite{RivaGAN} & 41.47 & 0.995 & 0.0370 & \textbf{0.99} & 0.88 & 0.86 & 0.96 & 0.98 & 0.98 \\ 
 & SepMark \cite{SepMark} & 35.24 & 0.939 & 0.0120 & 0.86 & 0.87 & 0.87 & 0.94 & 0.93 & 0.87   \\ 
 & WAM \cite{JNDWAM} & 42.25 & 0.994 & 0.0139 & \textbf{0.99} & 0.93 & 0.91 & 0.98 & 0.96 & \textbf{0.99}  \\ 
 & {\titleabbr } & \textbf{55.81} & \textbf{0.999} & \textbf{0.0009} & \textbf{0.99} & \textbf{0.99} & \textbf{0.99} & \textbf{0.99} & \textbf{0.99} & \textbf{0.99} \\ \hline

\multirow{7}{*}{\rotatebox{90}{{FaceDancer} \cite{facedancer2023}}} & DwtDct & 45.50 & 0.986 & 0.0013 & 0.75 & 0.74 & 0.74 & 0.74 & 0.74 & 0.75 \\ 
& DwtDctSvd \cite{DwtDctSvd} & 41.56 & 0.988 & 0.0046 & 0.85 & 0.74 & 0.74 & 0.75 & 0.81 & 0.80 \\ 
& MBRS \cite{MBRS} & 44.01 & 0.988 & 0.0024 & \textbf{0.99} & 0.81 & 0.89 & 0.87 & 0.87 & 0.98 \\ 
 & StegaStamp \cite{StegaStamp} & 35.34 & 0.969 & 0.0253 & \textbf{0.99} & \textbf{0.99} & \textbf{0.99} & 0.81 & 0.95 & \textbf{0.99} \\ 
 & RivaGAN \cite{RivaGAN} & 41.26 & 0.996 & 0.0380 & 0.98 & 0.87 & 0.85 & 0.94 & 0.98 & 0.98 \\
 & SepMark \cite{SepMark} & 35.21 & 0.932 & 0.0150 & 0.86 & 0.85 & 0.86 & 0.95 & 0.91 & 0.84  \\ 
 & WAM \cite{JNDWAM} & 41.69 & 0.994 & 0.0168 & \textbf{0.99} & 0.95 & 0.92 & \textbf{0.99} & 0.96 &  \textbf{0.99}  \\ 
 & {\titleabbr } & \textbf{56.16} & \textbf{0.999} & \textbf{0.0005} & \textbf{0.99} & \textbf{0.99} & \textbf{0.99} & 0.97 & \textbf{0.99} & \textbf{0.99} \\ \hline
 
\multirow{7}{*}{\rotatebox{90}{{Ghost} \cite{ghost2022}}} & DwtDct & 45.42 & 0.985 & 0.0015 & 0.75 & 0.74 & 0.74 & 0.74 & 0.73 & 0.75 \\ 
 & DwtDctSvd \cite{DwtDctSvd} & 41.61 & 0.988 & 0.0050 & 0.85 & 0.74 & 0.75 & 0.75 & 0.81 & 0.80 \\ 
 & MBRS \cite{MBRS} & 44.08 & 0.988 & 0.0025 & \textbf{0.99} & 0.81 & 0.89 & 0.87 & 0.86 & 0.98 \\ 
 & StegaStamp \cite{StegaStamp} & 35.30 & 0.969 & 0.0247 & \textbf{0.99} & \textbf{0.99} & \textbf{0.99} & 0.81 & 0.95 & \textbf{0.99} \\ 
 & RivaGAN \cite{RivaGAN} & 41.28 & 0.996 & 0.0380 & 0.98 & 0.86 & 0.85 & 0.95 & \textbf{0.98} & 0.98 \\ 
 & SepMark \cite{SepMark} & 35.13 & 0.929 & 0.0160 & 0.89 & 0.86 & 0.88 & 0.96 & 0.90 & 0.82  \\ 
 & WAM \cite{JNDWAM} & 41.75 & 0.994 & 0.0149 & \textbf{0.99} & 0.95 & 0.93 & \textbf{0.98} & 0.95 &  \textbf{0.99}  \\ 
 & {\titleabbr } & \textbf{57.49} & \textbf{0.999} & \textbf{0.0003} & \textbf{0.99} & 0.97 & \textbf{0.99} & \textbf{0.98} & \textbf{0.98} & \textbf{0.99} \\ \hline

\multirow{7}{*}{\rotatebox{90}{{SimSwap} \cite{simswap2020}}} & DwtDct & 45.46 & 0.986 & 0.0012 & 0.75 & 0.74 & 0.74 & 0.74 & 0.75 & 0.75 \\ 
 & DwtDctSvd \cite{DwtDctSvd} & 41.62 & 0.988 & 0.0042 & 0.85 & 0.75 & 0.75 & 0.75 & 0.82 & 0.80 \\ 
 & MBRS \cite{MBRS} & 43.81 & 0.989 & 0.0023 & \textbf{0.99} & 0.82 & 0.90 & 0.88 & 0.87 & 0.98 \\ 
 & StegaStamp \cite{StegaStamp} & 35.10 & 0.968 & 0.0256 & \textbf{0.99} & \textbf{0.99} & \textbf{0.99} & 0.81 & 0.95 & \textbf{0.99} \\ 
 & RivaGAN \cite{RivaGAN} & 41.21 & 0.996 & 0.0390 & 0.96 & 0.86 & 0.85 & 0.95 & 0.98 & 0.98 \\ 
 & SepMark \cite{SepMark} & 34.80 & 0.930 & 0.0150 & 0.91 & 0.91 & 0.88 & 0.94 & 0.90 & 0.87   \\
 & WAM \cite{JNDWAM} & 42.04 & 0.995 & 0.0154 & \textbf{0.99} & 0.96 & 0.94 & \textbf{0.98} & 0.96 & \textbf{0.99}   \\ 
 & {\titleabbr} & \textbf{51.40} & \textbf{0.998} & \textbf{0.0008} & \textbf{0.99} & \textbf{0.99} & \textbf{0.99} & \textbf{0.98} & \textbf{0.99} & \textbf{0.99} \\ \hline 

\multirow{7}{*}{\rotatebox{90}{{SimSwap++} \cite{simswap++2023}}} & DwtDct & 45.43 & 0.985 & 0.0012 & 0.74 & 0.74 & 0.74 & 0.74 & 0.74 & 0.75 \\ 
 & DwtDctSvd \cite{DwtDctSvd} & 41.62 & 0.987 & 0.0040 & 0.85 & 0.74 & 0.74 & 0.75 & 0.82 & 0.80 \\ 
 & MBRS \cite{MBRS} & 43.83 & 0.988 & 0.0025 & \textbf{0.99} & 0.81 & 0.89 & 0.87 & 0.87 & 0.97 \\ 
 & StegaStamp \cite{StegaStamp} & 35.14 & 0.967 & 0.0254 & \textbf{0.99} & \textbf{0.99} & \textbf{0.99} & 0.81 & 0.95 & \textbf{0.99} \\ 
 & RivaGAN \cite{RivaGAN} & 41.27 & 0.996 & 0.0380 & \textbf{0.99} & 0.89 & 0.87 & 0.96 & 0.98 & 0.98 \\ 
 & SepMark \cite{SepMark} & 34.91 & 0.932 & 0.0140 & 0.88 & 0.86 & 0.86 & 0.96 & 0.91 & 0.82 \\
 & WAM \cite{JNDWAM} & 42.03 & 0.995 & 0.0143 & \textbf{0.99} & 0.95 & 0.95 & \textbf{0.98} & 0.96 & \textbf{0.99}  \\ 
 & {\titleabbr} & \textbf{53.21} & \textbf{0.999} & \textbf{0.0006} & \textbf{0.99} & \textbf{0.99} & \textbf{0.99} & 0.97 & \textbf{0.99} & \textbf{0.99} \\ \hline

\multirow{7}{*}{\rotatebox{90}{{InsightFace} \cite{InsightFace}}} & DwtDct & 45.55 & 0.986 & 0.0015 & 0.75 & 0.74 & 0.74 & 0.74 & 0.74 & 0.75 \\ 
 & DwtDctSvd \cite{DwtDctSvd} & 41.67 & 0.988 & 0.0046 & 0.85 & 0.75 & 0.75 & 0.75 & 0.82 & 0.80 \\ 
 & MBRS \cite{MBRS} & 43.98 & 0.98 & 0.0024 & \textbf{0.99} & 0.83 & 0.90 & 0.87 & 0.87 & 0.98 \\ 
 & StegaStamp \cite{StegaStamp} & 35.26 & 0.967 & 0.0256 & \textbf{0.99} & \textbf{0.99} & \textbf{0.99} & 0.81 & 0.95 & \textbf{0.99} \\ 
 & RivaGAN \cite{RivaGAN} & 41.27 & 0.996 & 0.0376 & 0.98 & 0.87 & 0.86 & 0.95 & 0.98 & 0.98 \\
 & SepMark \cite{SepMark} & 35.15 & 0.927 & 0.0160 & 0.91 & 0.87 & 0.88 & 0.96 & 0.91 & 0.85 \\
 & WAM \cite{JNDWAM} & 41.82 & 0.992 & 0.0138 & \textbf{0.99} & 0.96 & 0.95 & \textbf{0.99} & 0.97 & \textbf{0.99}  \\ 
 & {\titleabbr} & \textbf{51.30} & \textbf{0.998} & \textbf{0.0005} & \textbf{0.99} & \textbf{0.99} & \textbf{0.99} & \textbf{0.99} & \textbf{0.99} & \textbf{0.99} \\ \hline
\end{tabular}
\end{adjustbox}
\end{table*}

%% file: Table/table_2.tex
\begin{table}[!ht]
\scriptsize 
\centering
\caption{Cross-dataset robustness evaluation on public datasets}
\label{Public}
\begin{adjustbox}{width=0.4\textwidth}
\begin{tabular}{l  l   c|  c|c|c|c}
\hline
\multicolumn{4}{l |}{\textbf{Dataset}} & \textbf{CelebA} \cite{liu2015faceattributes} & \textbf{FF++} \cite{rossler2019faceforensics++} & \textbf{FF} \cite{rossler2018faceforensics} \\
\hline \hline 

\multicolumn{3}{l|}{\multirow{3}{*}{{\textbf{Imperceptibility}} }} 
& PSNR $\uparrow$ & 55.89 & 55.92 & 55.83 \\
&& & SSIM $\uparrow$ & 0.9994 & 0.9996 & 0.9993 \\
&& & LPIPS $\downarrow$ & 0.0016 & 0.0007 & 0.0019 \\
\hline \hline 

\multicolumn{3}{l|}{\multirow{2}{*}{\textbf{Pre-Attack}}} 
& $\mathcal{A}_{ssim}$$\uparrow$ & 0.99 & 0.99 & 0.99 \\ 
\multicolumn{3}{l|}{} 
& $\mathcal{A}_{br}$ $\uparrow$ & 0.99 & 0.99 & 0.99 \\
\hline \hline 

\multirow{16}{*}{\rotatebox{90}{\textbf{Post-Attack}}}  

& \multicolumn{1}{|l|}{\multirow{8}{*}{\rotatebox{90}{\scriptsize Instagram Filters}}}
& \multirow{2}{*}{$\mathbb{A}$} 
& $\mathcal{A}_{ssim}$$\uparrow$ & 0.99 & 0.99 & 0.99 \\
& \multicolumn{1}{|l|}{} 
& 
& $\mathcal{A}_{br}$$\uparrow$ & 0.99 & 0.99 & 0.99 \\
\cline{3-7}

& \multicolumn{1}{|l|}{} 
& \multirow{2}{*}{$\mathbb{B}$} 
& $\mathcal{A}_{ssim}$$\uparrow$ & 0.99 & 0.99 & 0.99 \\
& \multicolumn{1}{|l|}{} 
& 
& $\mathcal{A}_{br}$$\uparrow$ & 0.99 & 0.99 & 0.99 \\
\cline{3-7}

& \multicolumn{1}{|l|}{} 
& \multirow{2}{*}{$\mathbb{C}$} 
& $\mathcal{A}_{ssim}$$\uparrow$ & 0.99 & 0.99 & 0.99 \\
& \multicolumn{1}{|l|}{} 
& 
& $\mathcal{A}_{br}$$\uparrow$ & 0.99 & 0.99 & 0.99 \\
\cline{3-7}

& \multicolumn{1}{|l|}{} 
& \multirow{2}{*}{$\mathbb{A}{+}\mathbb{B}{+}\mathbb{C}$} 
& $\mathcal{A}_{ssim}$$\uparrow$ & 0.99 & 0.99 & 0.99 \\
& \multicolumn{1}{|l|}{} 
& 
& $\mathcal{A}_{br}$$\uparrow$ & 0.99 & 0.99 & 0.99 \\

\cline{2-7}

& \multicolumn{1}{|l|}{\multirow{4}{*}{\rotatebox{90}{JPEG}}}
& \multirow{2}{*}{$\text{QF}=50\%$} 
& $\mathcal{A}_{ssim}$$\uparrow$ & 0.93 & 0.96 & 0.97 \\
& \multicolumn{1}{|l|}{} 
& 
& $\mathcal{A}_{br}$$\uparrow$ & 0.96 & 0.98 & 0.98 \\
\cline{3-7}

& \multicolumn{1}{|l|}{} 
& \multirow{2}{*}{$\text{QF}=75\%$} 
& $\mathcal{A}_{ssim}$$\uparrow$ & 0.99 & 0.99 & 0.99 \\
& \multicolumn{1}{|l|}{} 
& 
& $\mathcal{A}_{br}$$\uparrow$ & 0.99 & 0.99 & 0.99 \\

\cline{2-7}

& \multicolumn{1}{|l|}{\multirow{4}{*}{\rotatebox{90}{G-Blur}}}
& \multirow{2}{*}{$\mathcal{K}=3{\times}3$} 
& $\mathcal{A}_{ssim}$$\uparrow$ & 0.99 & 0.99 & 0.99 \\
& \multicolumn{1}{|l|}{} 
& 
& $\mathcal{A}_{br}$$\uparrow$ & 0.99 & 0.99 & 0.99 \\
\cline{3-7}

& \multicolumn{1}{|l|}{} 
& \multirow{2}{*}{$\mathcal{K}=5{\times}5$} 
& $\mathcal{A}_{ssim}$$\uparrow$ & 0.99 & 0.99 & 0.99 \\
& \multicolumn{1}{|l|}{} 
& 
& $\mathcal{A}_{br}$$\uparrow$ & 0.99 & 0.99 & 0.99 \\

\hline

\multicolumn{7}{r}{\scriptsize Instagram filters: $\mathbb{A}$ (Aden), $\mathbb{B}$ (Brooklyn), $\mathbb{C}$ (Clarendon)}

\end{tabular}
\end{adjustbox}
\end{table}

%% file: Image/Fig_2_Qualitative_Analysis/Q_Analysis.tex
\begin{figure*}[!t]
\centering
\footnotesize
\setlength{\tabcolsep}{2.5pt}
\newcolumntype{C}[1]{>{\centering\arraybackslash}m{#1}}

\begin{adjustbox}{width=0.65\textwidth}
\begin{tabular}{C{0.9cm} | C{1.8cm} |C{1.8cm} |C{1.8cm} |C{1.8cm} |C{1.8cm} |C{1.8cm} |C{1.8cm} |C{1.8cm}}
\cline{2-9}
\multicolumn{1}{c}{}
& {Input image ($\mathcal{I}$)} 
& {Watermark logo ($\mathcal{I}_l$)} 
& {Perceptual map ($\mathcal{P}_n$)} 
& {Bottleneck feature ($g_h'$)} 
& {Residual map ($\mathcal{I}_r$)} 
& {Watermarked image ($\mathcal{I}_w$)} 
& {Distorted image ($\mathcal{I}_d$)} 
& {Recovered logo ($\hat{\mathcal{I}}_l$)} \\

\hline &&&& &&&& \\ [\dimexpr-\normalbaselineskip+1.5pt]
\rotatebox{90}{{FaceApp}} &
\includegraphics[width=0.88\linewidth]{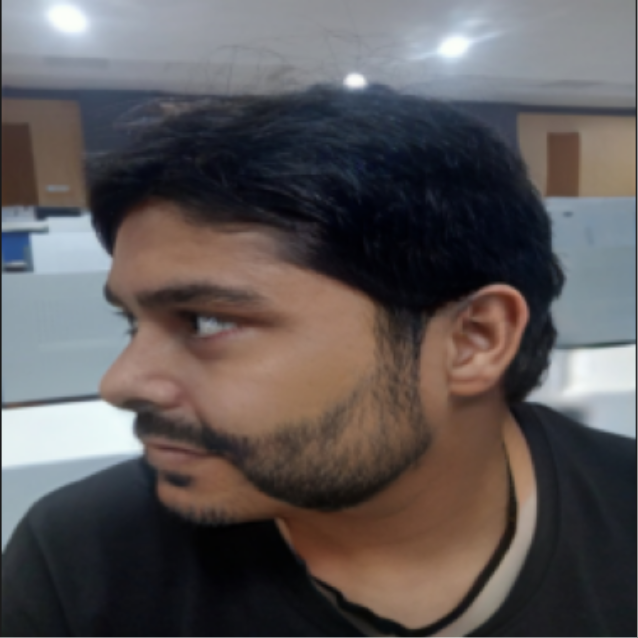} &
\includegraphics[width=0.62\linewidth]{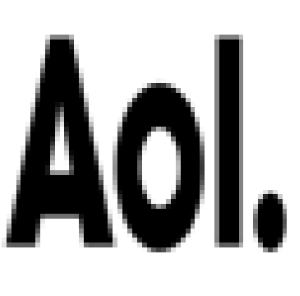} &
\includegraphics[width=0.88\linewidth]{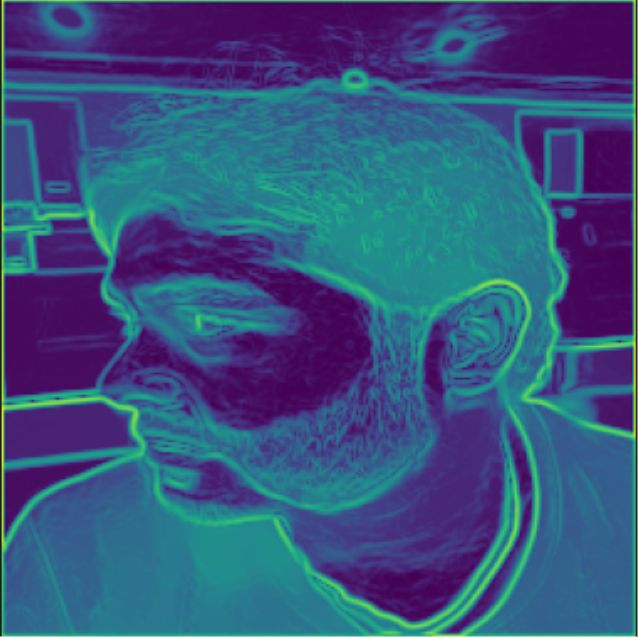} &
\includegraphics[width=0.88\linewidth]{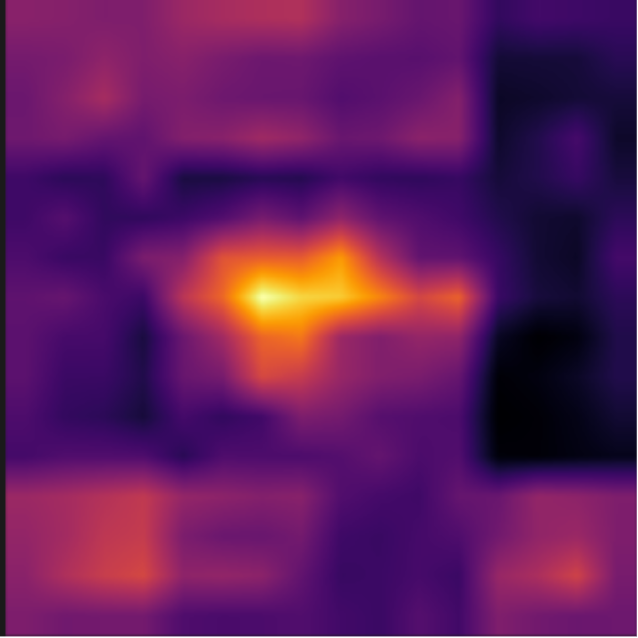} &
\includegraphics[width=0.88\linewidth]{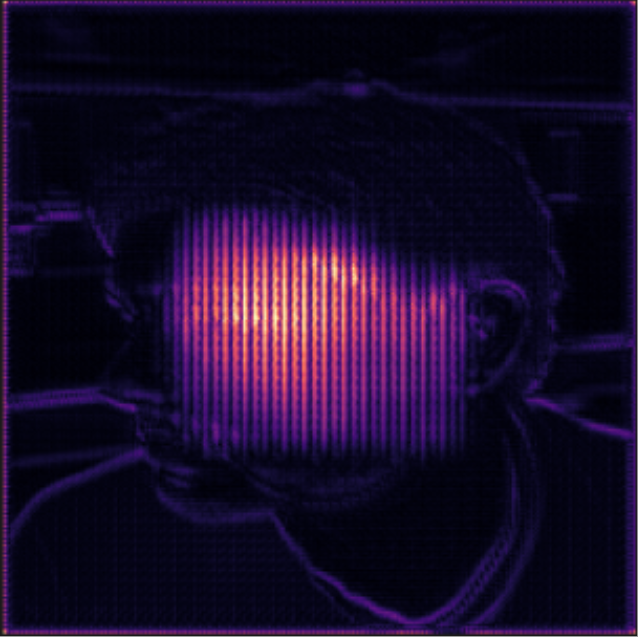} &
\includegraphics[width=0.88\linewidth]{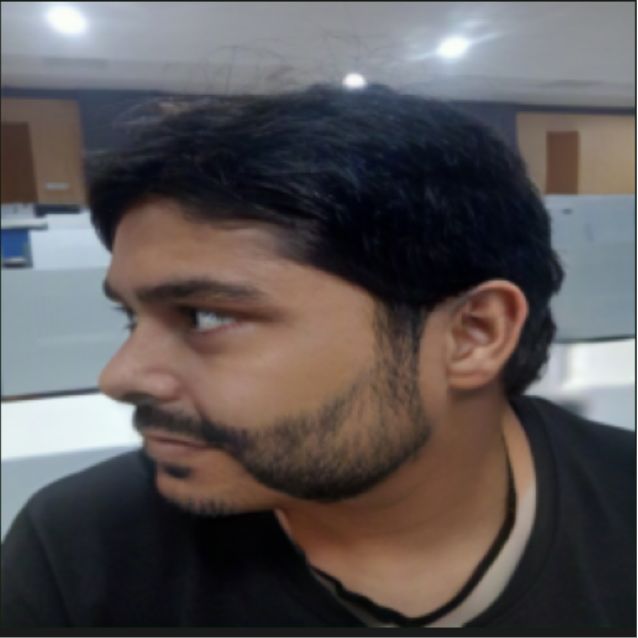} &
\includegraphics[width=0.88\linewidth]{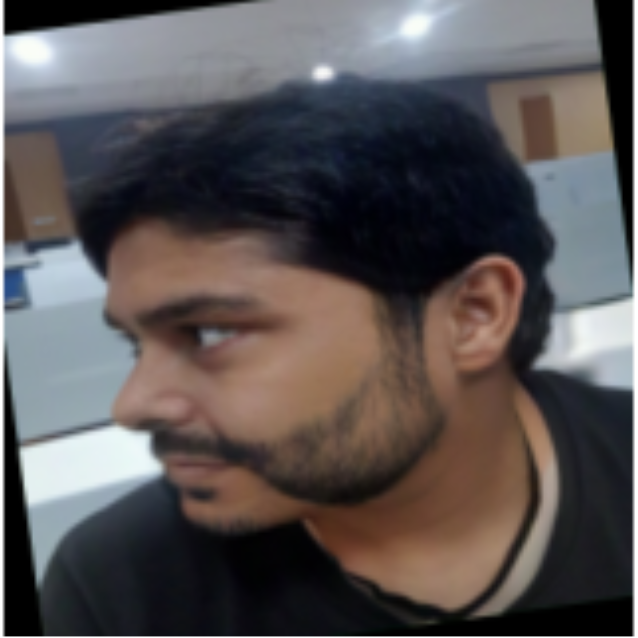} &
\includegraphics[width=0.62\linewidth]{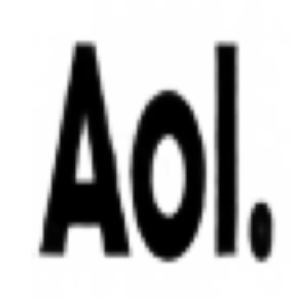} \\
  
\hline &&&& &&&& \\ [\dimexpr-\normalbaselineskip+1.5pt]
\rotatebox{90}{{FaceDancer}} & 
\includegraphics[width=0.88\linewidth]{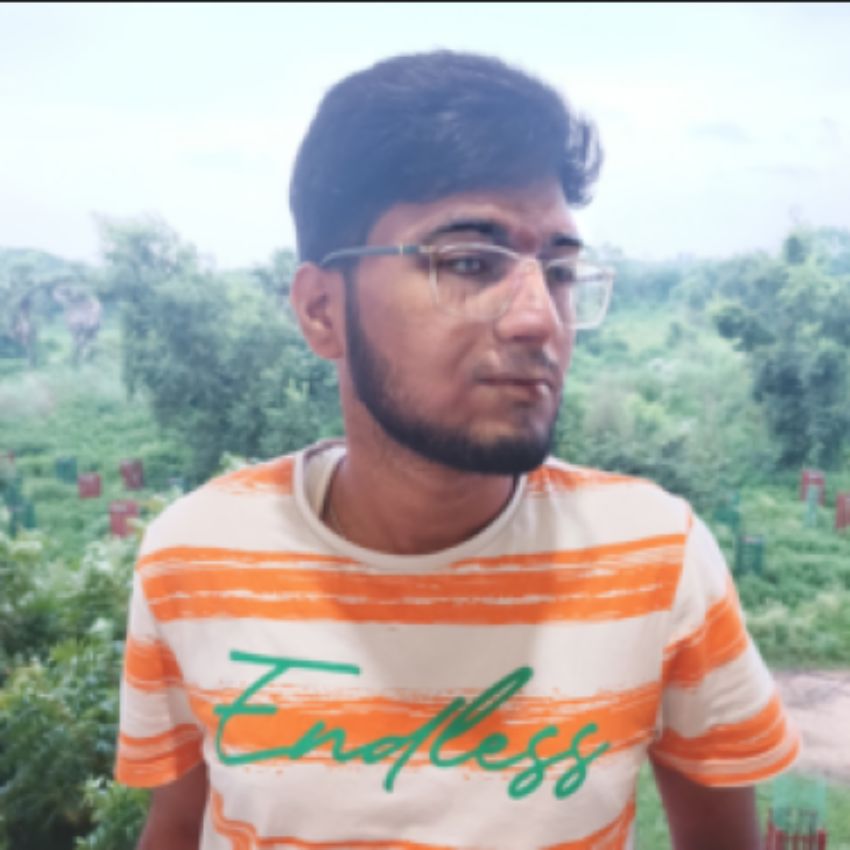} &
\includegraphics[width=0.62\linewidth]{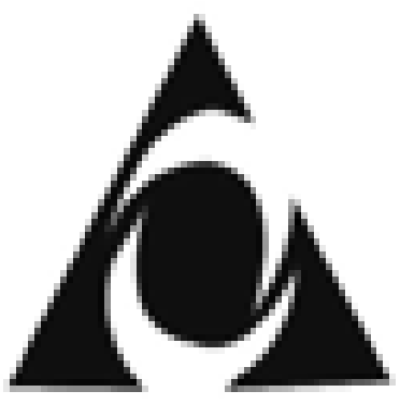} &
\includegraphics[width=0.88\linewidth]{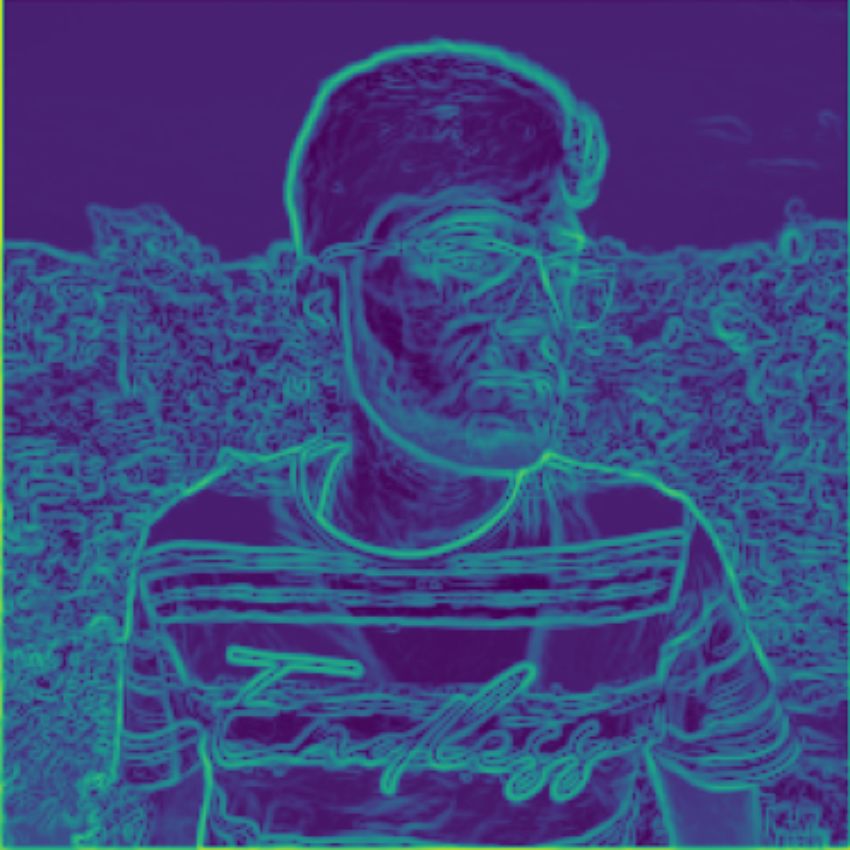} &
\includegraphics[width=0.88\linewidth]{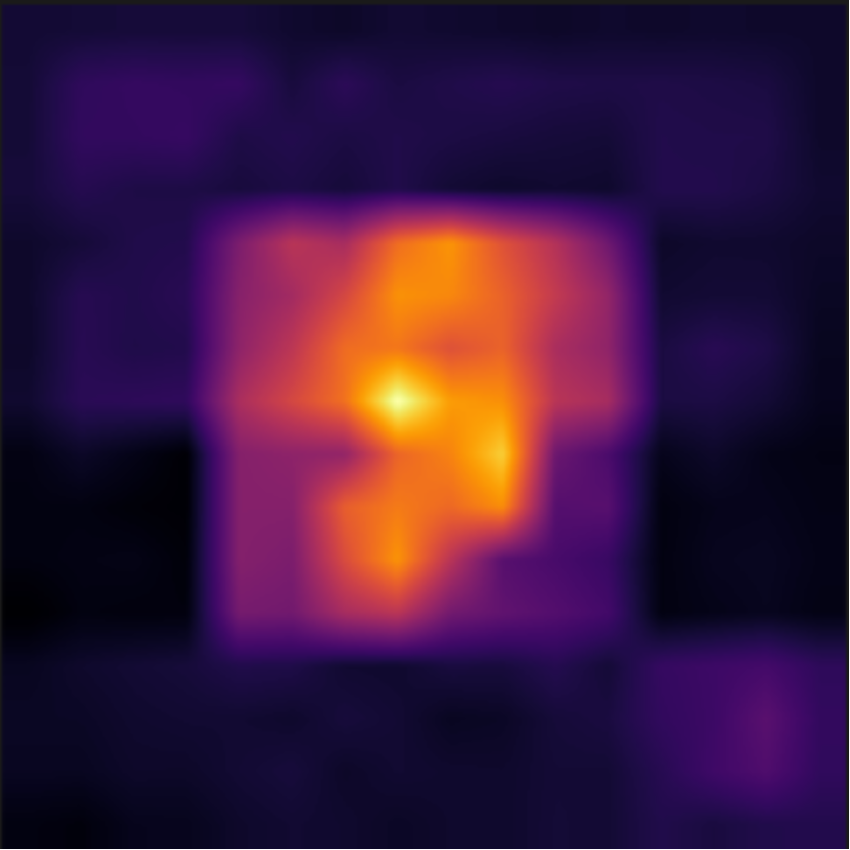} &
\includegraphics[width=0.88\linewidth]{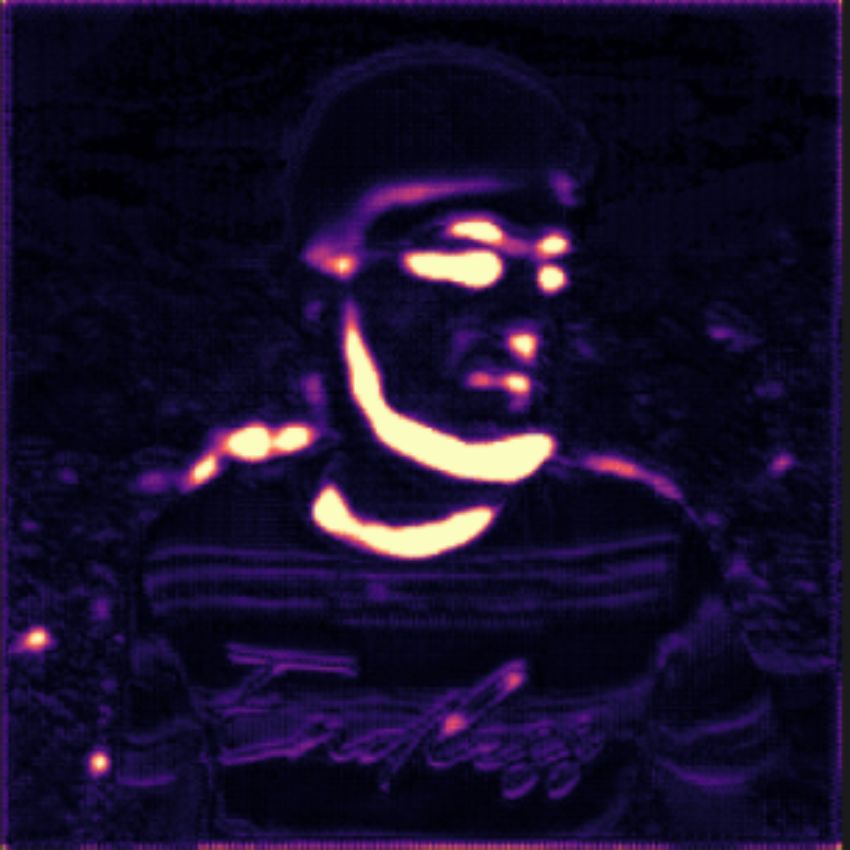} &
\includegraphics[width=0.88\linewidth]{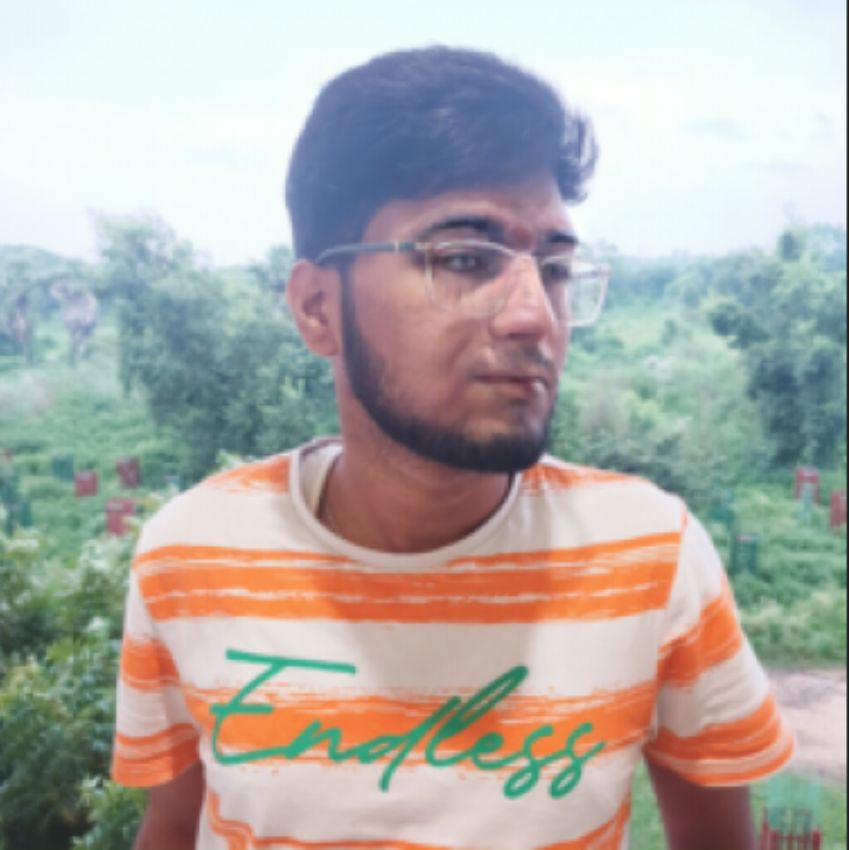} &
\includegraphics[width=0.88\linewidth]{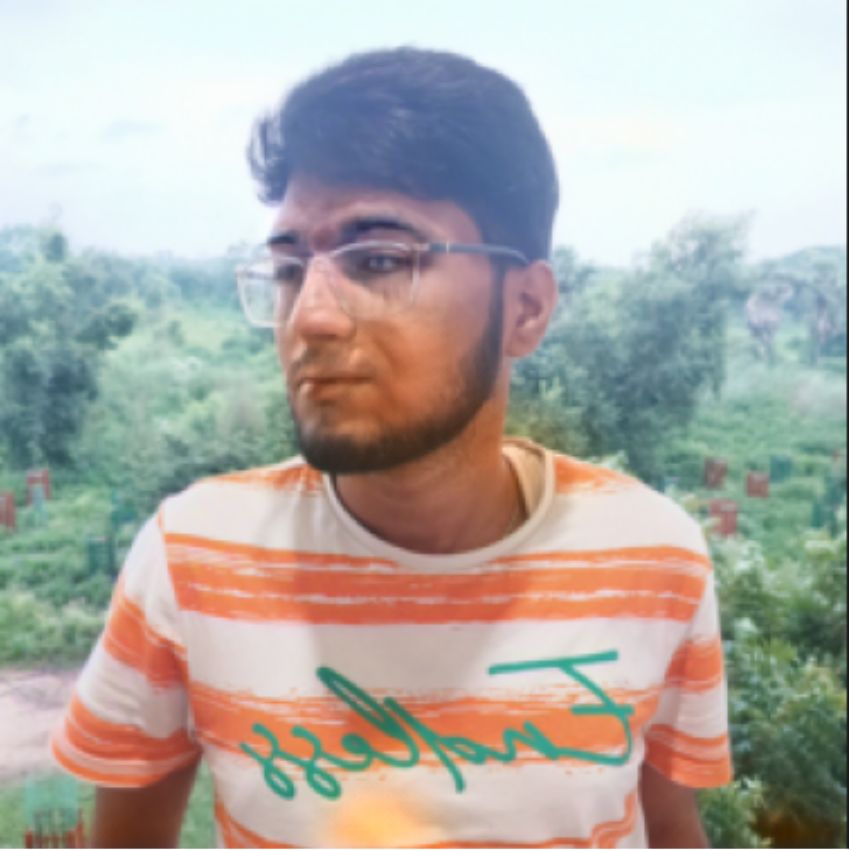} &
\includegraphics[width=0.62\linewidth]{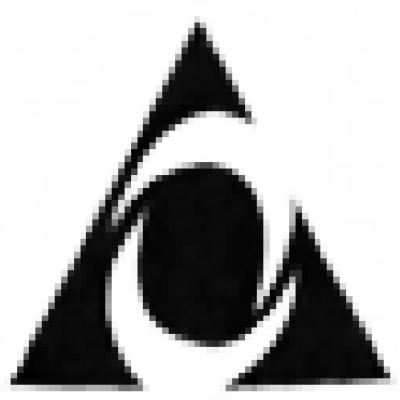} \\
  
\hline &&&& &&&& \\ [\dimexpr-\normalbaselineskip+1.5pt]
\rotatebox{90}{{Ghost}} &
\includegraphics[width=0.88\linewidth]{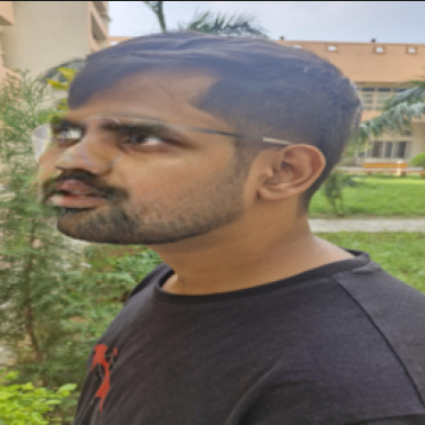} &
\includegraphics[width=0.62\linewidth]{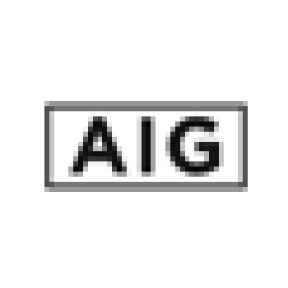} &
\includegraphics[width=0.88\linewidth]{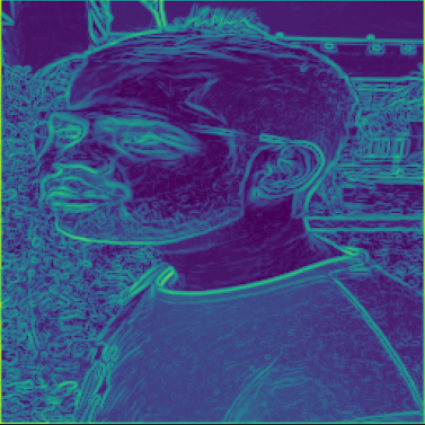} &
\includegraphics[width=0.88\linewidth]{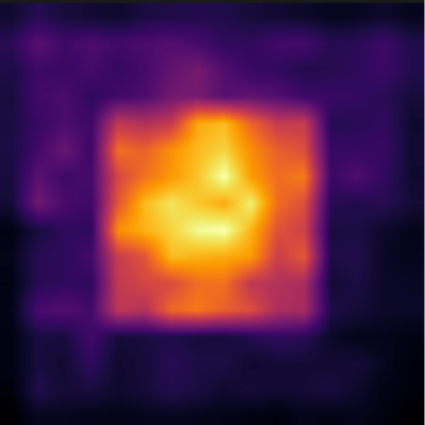} &
\includegraphics[width=0.88\linewidth]{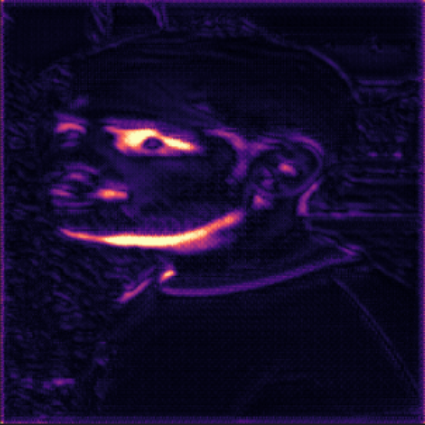} &
\includegraphics[width=0.88\linewidth]{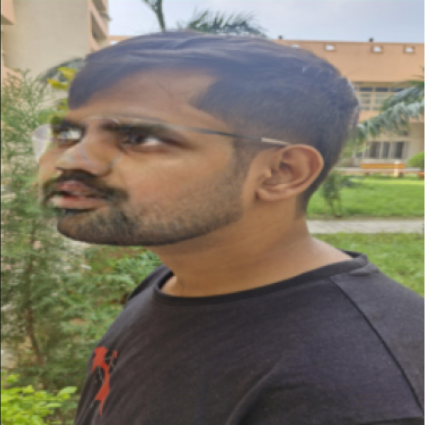} &
\includegraphics[width=0.88\linewidth]{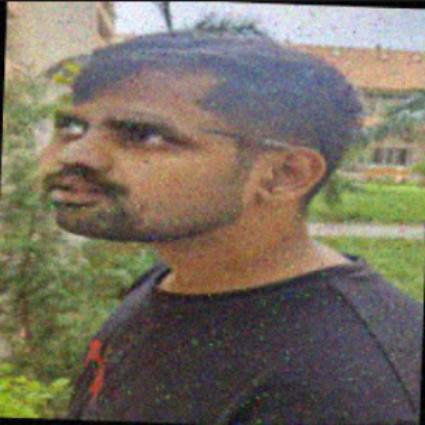} &
\includegraphics[width=0.62\linewidth]{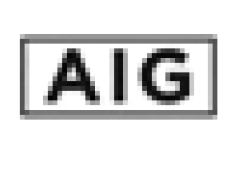} \\
  
\hline &&&& &&&& \\ [\dimexpr-\normalbaselineskip+1.5pt]
\rotatebox{90}{{SimSwap}} &
\includegraphics[width=0.88\linewidth]{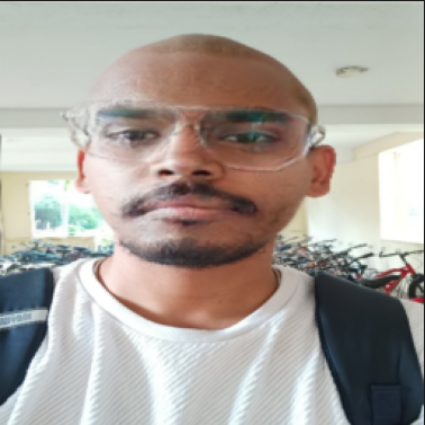} &
\includegraphics[width=0.62\linewidth]{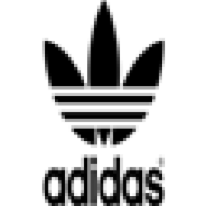} &
\includegraphics[width=0.88\linewidth]{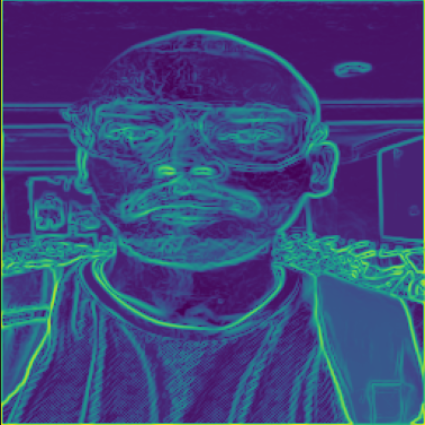} &
\includegraphics[width=0.88\linewidth]{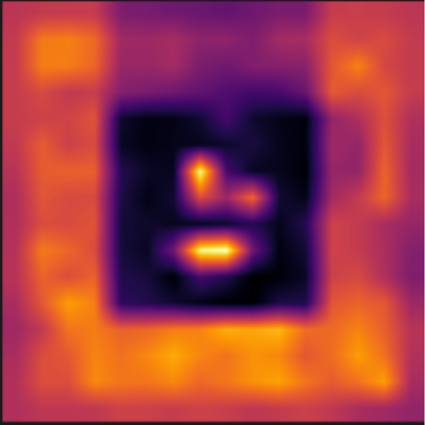} &
\includegraphics[width=0.88\linewidth]{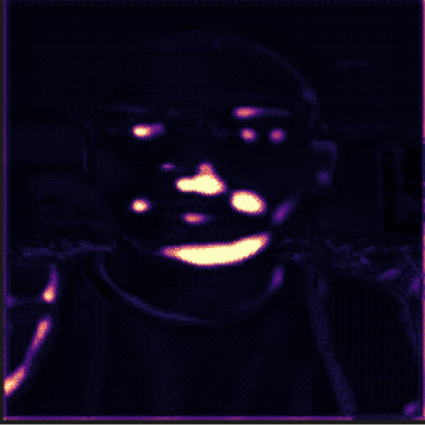} &
\includegraphics[width=0.88\linewidth]{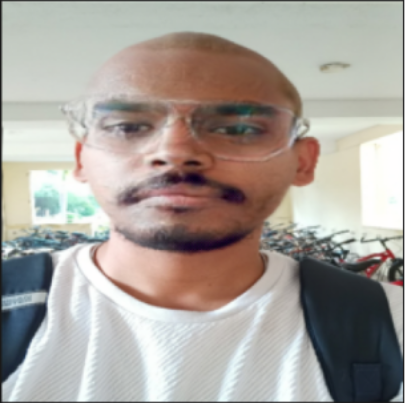} &
\includegraphics[width=0.88\linewidth]{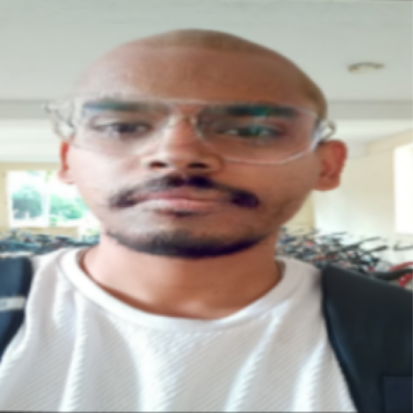} &
\includegraphics[width=0.62\linewidth]{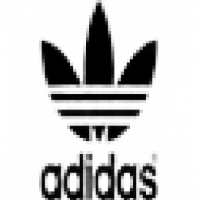} \\
  
\hline &&&& &&&& \\ [\dimexpr-\normalbaselineskip+1.5pt]
\rotatebox{90}{{InsightFace}} &
\includegraphics[width=0.88\linewidth]{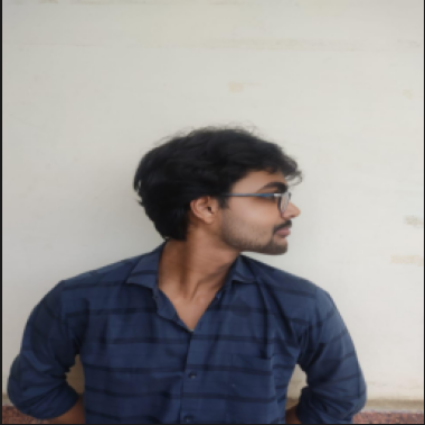} &
\includegraphics[width=0.62\linewidth]{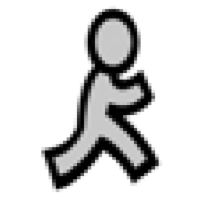} &
\includegraphics[width=0.88\linewidth]{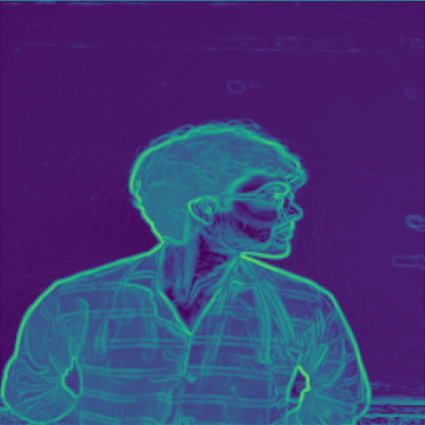} &
\includegraphics[width=0.88\linewidth]{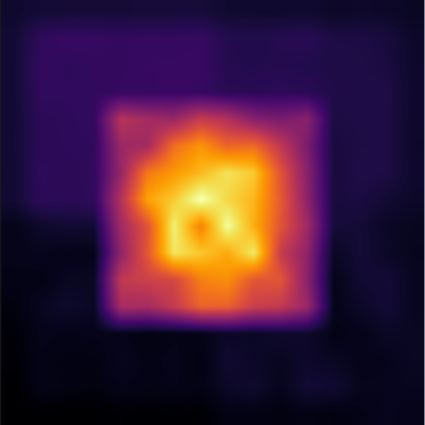} &
\includegraphics[width=0.88\linewidth]{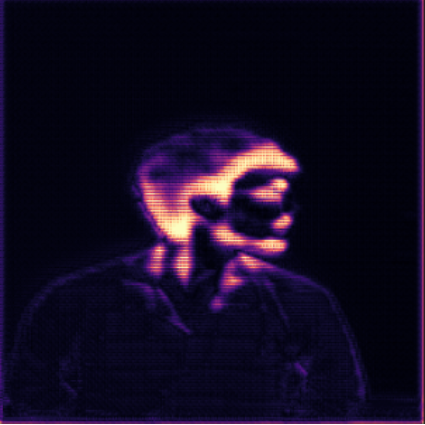} &
\includegraphics[width=0.88\linewidth]{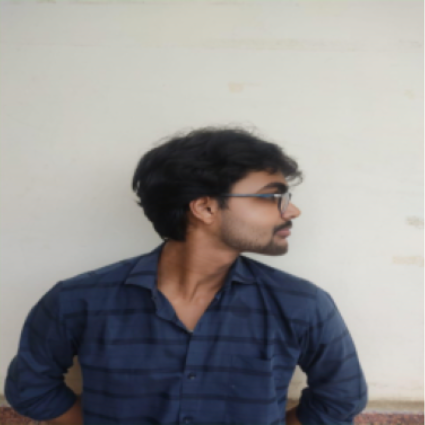} &
\includegraphics[width=0.88\linewidth]{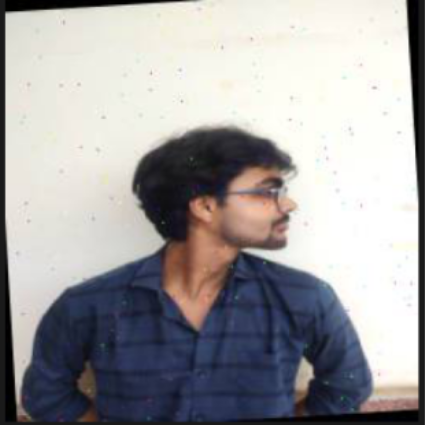} &
\includegraphics[width=0.62\linewidth]{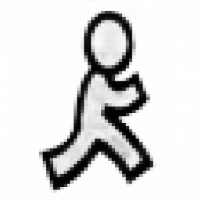} \\
  
\hline &&&& &&&& \\ [\dimexpr-\normalbaselineskip+1.5pt]
\rotatebox{90}{{SimSwap++}} &
\includegraphics[width=0.88\linewidth]{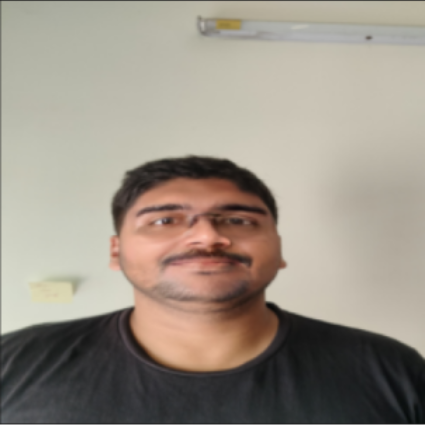} &
\includegraphics[width=0.62\linewidth]{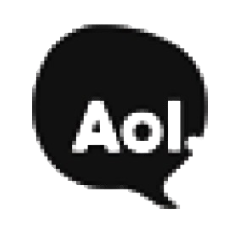} &
\includegraphics[width=0.88\linewidth]{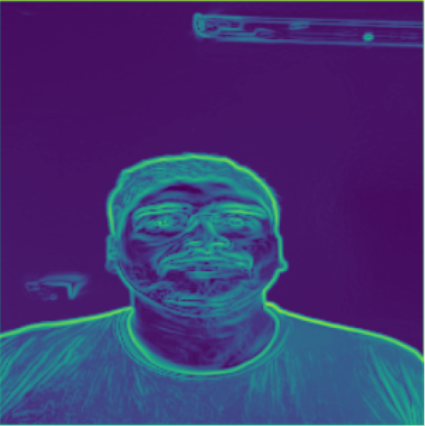} &
\includegraphics[width=0.88\linewidth]{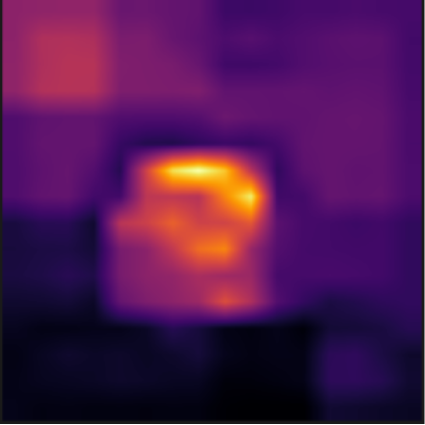} &
\includegraphics[width=0.88\linewidth]{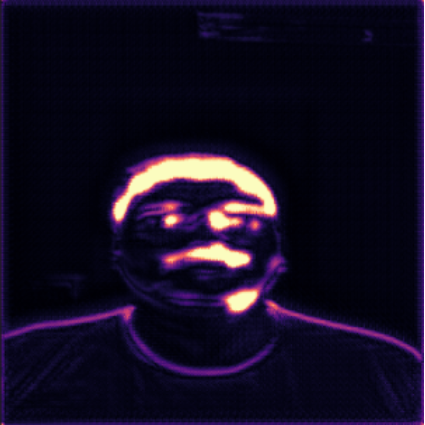} &
\includegraphics[width=0.88\linewidth]{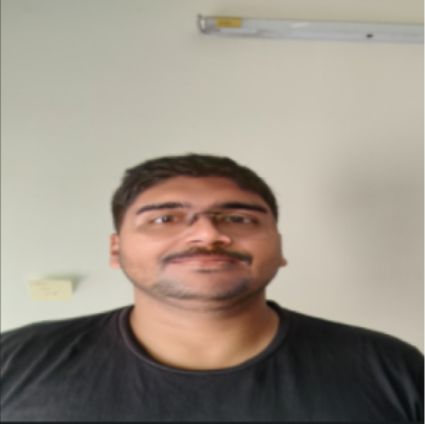} &
\includegraphics[width=0.88\linewidth]{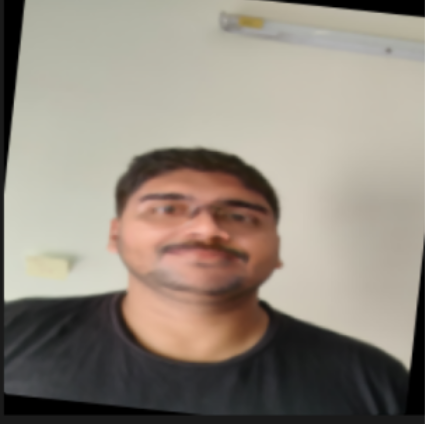} &
\includegraphics[width=0.62\linewidth]{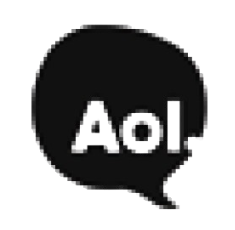} \\
  
\hline &&&& &&&& \\ [\dimexpr-\normalbaselineskip+1.5pt]
\rotatebox{90}{{Genuine}} &
\includegraphics[width=0.88\linewidth]{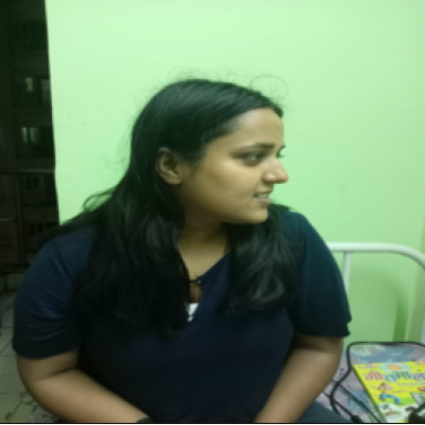} &
\includegraphics[width=0.62\linewidth]{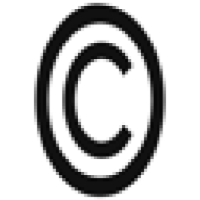} &
\includegraphics[width=0.88\linewidth]{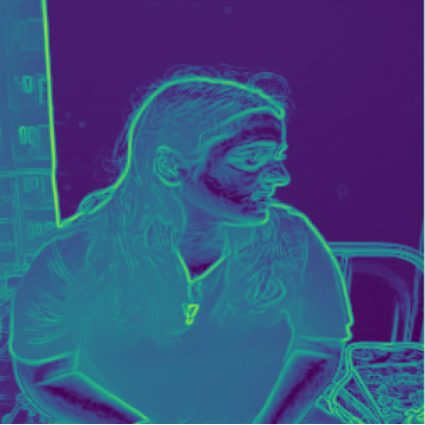} &
\includegraphics[width=0.88\linewidth]{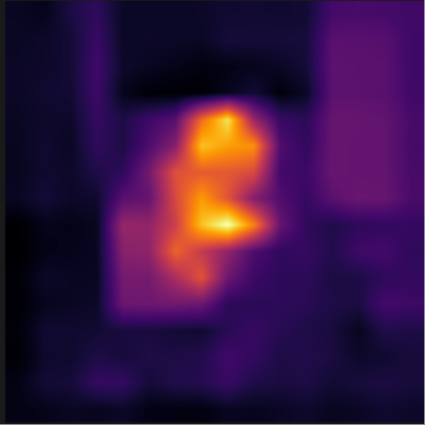} &
\includegraphics[width=0.88\linewidth]{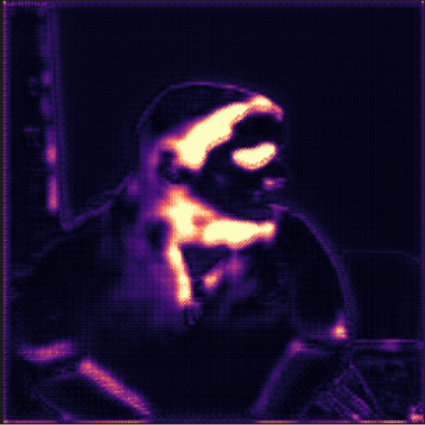} &
\includegraphics[width=0.88\linewidth]{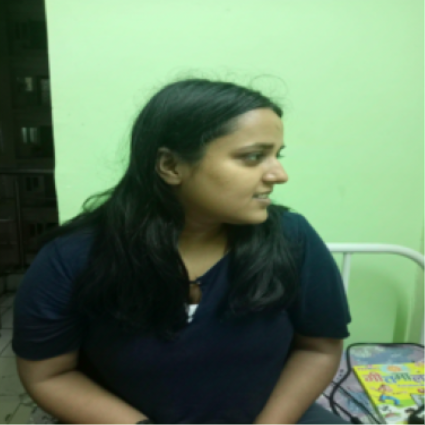} &
\includegraphics[width=0.88\linewidth]{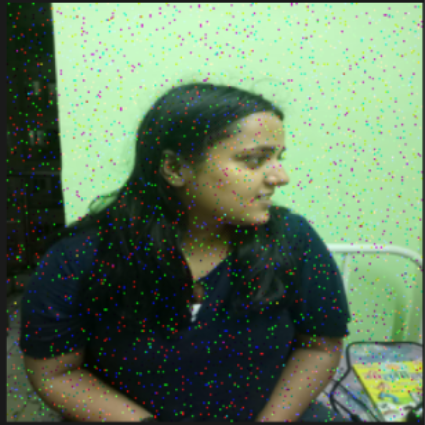} &
\includegraphics[width=0.62\linewidth]{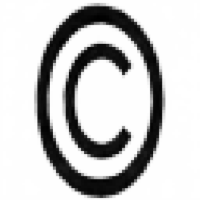} \\ \hline 
 
\end{tabular}
\end{adjustbox}

\caption{\small Qualitative analysis of the proposed framework {\titleabbr} on IndicSideFace \cite{IndicSideFace2025}. Softcopy exhibits better display.}
\label{Q1}
\end{figure*}



%% file: 6Conclusion.tex
\section{Conclusion}
\label{conclusion}

This work introduces {\titleabbr}, a source-conditioned deep watermarking framework for proactive media authentication and provenance verification. Unlike conventional watermarking methods that embed generic payloads, the proposed formulation treats watermark identity as a conditioning signal that directly guides the embedding process, enabling a unified model to learn source-aware watermark representations. By integrating a hybrid convolution–transformer embedding network with perceptual guidance inspired by human visual system principles, the framework distributes watermark perturbations adaptively while maintaining visual fidelity. Furthermore, the proposed dual-purpose forensic decoder jointly performs watermark reconstruction and source attribution, providing both automated verification and interpretable provenance evidence. Through this design, {\titleabbr} establishes a scalable and reliable mechanism for binding digital media to its origin, offering a practical foundation for trustworthy media provenance and strengthening proactive defenses against synthetic media manipulation. 
Future work will extend the framework to video and multimodal media and explore robustness against generative regeneration pipelines to support large-scale provenance tracking in evolving generative ecosystems.